\documentclass[journal,twoside]{IEEEtran}
\usepackage{lineno}
\usepackage{amsmath,array}
\usepackage{subfigure}
\usepackage{multirow}
\usepackage[ruled]{algorithm2e}
\usepackage{graphicx}

\begin{document}


\title{ASBSO: An Improved Brain Storm Optimization with Flexible Search Length and Memory-based Selection}


\author{Yang Yu, Shangce Gao, \IEEEmembership{Senior Member, IEEE}, Yirui Wang, Jiujun Cheng and Yuki Todo, \IEEEmembership{Member, IEEE}
\thanks{\quad This research was partially supported by the National Natural Science Foundation of China (Grant Nos. 61472284, 61673403) and JSPS KAKENHI Grant Number JP17K12751, JP15K00332. (Corresponding authors: Shangce Gao (gaosc@eng.u-toyama.ac.jp), Jiujun Cheng (chengjj@tongji.edu.cn))}
\thanks{\quad Yang Yu, Shangce Gao and Yirui Wang 
        are with the Faculty of Engineering, University of Toyama, Toyama 930-8555, Japan.
        (e-mail: gaosc@eng.u-toyama.ac.jp).}
\thanks{\quad Jiujun Cheng is with Key Laboratory of Embedded System and Service Computing, Ministry of Education, Department of Computer Science and Technology, Tongji University, Shanghai, 200092 China.}
\thanks{\quad Yuki Todo is with the School of Electrical and Computer Engineering, Kanazawa University, Kanazawa-shi, 920-1192 Japan.}
}
\markboth{IEEE Access}{}

\maketitle

\begin{abstract}
Brain storm optimization (BSO) is a newly proposed population-based optimization algorithm which uses a logarithmic sigmoid transfer function to adjust its search range during the convergent process. However, this adjustment only varies with the current iteration number, and lacks of flexibility and variety which makes a poor search efficiency and robustness of BSO. To alleviate this problem, an adaptive step length structure together with a success memory selection strategy are proposed to be incorporated into BSO. This proposed method, adaptive step length based on memory selection BSO, namely ASBSO, applies multiple step lengths to modify the generation process of new solutions, thus supplying a flexible search according to corresponding problems and convergent periods. The novel memory mechanism which is capable of evaluating and storing the degree of improvements of solutions is used to determine the selection possibility of step lengths. A set of 57 benchmark functions are used to test ASBSO's search ability, and four real-world problems are adopted to show its application value. All these test results indicate the remarkable improvement in solution quality, scalability and robustness of ASBSO.
\end{abstract}


\IEEEpeerreviewmaketitle

\section{Introduction}
\IEEEPARstart{N}owadays, many swarm intelligence algorithms have been proposed to solve complex real-world problems \cite{yang2016hybrid, gao2016ant}.Brain storm optimization algorithm (BSO) which is one of the swarm intelligence algorithm, is promising in solving complex problems \cite{shi2011brain}. It is inspired by the human brain storming behaviors. Each idea generated by the human brain represents an individual in search space. In a brain storming process, humans firstly generate some rough ideas, then exchange and discuss these ideas with each other. The inferior ideas are sifted out while the superior ones are left. This operation circles over and over, which makes ideas become more and more mature. In the meanwhile, new ideas are kept being generated and joined in the circle. With the process ends, a feasible and effective idea spurts out.

Since the announcement of BSO in 2011, it gets lots of attention from the researchers in swarm intelligence community due to its novelty and efficiency.  It has been successfully applied in different scenarios, such as function optimization, engineering problems and financial prediction \cite{guo2014modified, guo2015adaptive, li2015history, qiu2014receding, sun2014hybrid}. Moreover, some modifications for BSO have been made to enhance its performance from several perspectives. For example, a new multi-objective BSO (MBSO) is proposed in \cite{shi2013multi} for solving multi-objective optimization problems. The clustering strategy is applied in the objective search space to handle multi-objective optimization problems, while it is originally performed in the solution search space for solving single objective problems. With different characteristics of diverging operation, MBSO becomes a promising algorithm with an outstanding ability to solve multi-objective optimization problems. In \cite{shi2015brain}, BSO in objective space (BSOOS) is proposed to cut down the computation time of the convergent operation. A clustering operation is replaced by taking $p$ percentage individuals as elitists. An updating operation is modified to suit for an elitists mechanism in one-dimensional objective space instead of solution space. By doing so, BSOOS achieves a better convergent speed and solution quality in comparison with the traditional BSO.

Improving the population diversity is an alternative modification besides the usage of objective space. As the balance between convergence and divergence is very important to swarm intelligence optimization algorithms, a premature convergence leads to a low population diversity and bad solution quality, while the opposite brings very slow search speed. The issue of how to find the balance between convergence and divergence of solutions is still very challenging and it reflects the algorithm's exploration and exploitation ability. In \cite{li2015information, yu2017cbso}, chaotic sequences are used as variables to initialize population and generate new individuals. As a universal phenomenon of nonlinear dynamic systems, chaos has an unpredictable random behavior \cite{gao2014gravitational}. Thus, its randomicity and ergodicity can help BSO improve its population diversity and solution quality effectively. In \cite{cheng2014population}, Cheng et al. propose a new BSO which uses different kinds of partial reinitialization strategies to increase its population diversity. Duan et al. \cite{duan2013predator} propose a novel predator-prey model to improve the population diversity of BSO for a DC brush-less motor. This model can enable the algorithm structure to explore the search space more evenly. By using the predator-prey strategy, the population can share better global information with each other to improve search efficiency in exploitation phase. In \cite{duan2015quantum}, quantum-behaved BSO (QBSO) which aims to improve population diversity and generate new individuals by using global information is proposed. Moreover, QBSO for the first time combines BSO with quantum theories. It analyzes the quantum behavior and quantum state of each individual by depicting a wave function to solve the drawback of BSO that easily sticks into local optima on multimodal functions. In addition, Wang et al. \cite{wang2017discovery} discover a power law distribution in BSO which opens a new way of thinking to boost the population interaction and improve population diversity via adjusting the population structure.

Although above mentioned modifications have improved the performance of BSO, they are limited and the performance of BSO is still fatigued and week \cite{cheng2016brain}. Most efforts attempt to modify BSO for solving specific problems while these modifications are not suitable for other applications. It is still a great demand to enhance its search ability and robustness.

To achieve this goal, we propose an adaptive step length mechanism based on memory selection to combine with BSO (namely, ASBSO) which exhibits a notable performance. This method can modify BSO by providing strategies with various step lengths which are adaptively applied to generate new individuals.  As it can supply a specific step length according to corresponding problems and convergent periods, it is more possible that ASBSO can avoid or jump out of the local optima. In other words, the search efficiency and robustness of BSO can be greatly improved.

Besides the adaptive step length mechanism, a modified selection method is also proposed based on memory. Different from the conventional storage mode in \cite{song2017multiple} which applies a success memory and a failure memory with 0 and 1 as the information stored in these memories, the modified method only employs the success memory and considers the difference between two compared fitness values instead of simple numbers (i.e., 0 and 1). This is a modification which directly demonstrates the improvement of each selected strategy and extrudes a strategy with a better performance. A detailed description is presented in Section \ref{section3}.

The contributions of this paper can be summarized as: (1) An adaptive step length mechanism based on memory selection method is proposed to enhance the robustness of BSO evidently, therefore makes it more suitable for various applications. (2) It's for the first time that we use the difference between two compared fitness values instead of simple numbers such as 0 and 1 to be stored in memory. This modification can increase the efficiency of the selection method, and thereby improve solution quality observably. An experimental comparison between the new storage mode and the old one brings an intuitional conclusion that the proposed method is significantly better. (3) Sufficient experimental data and statistical analyses of performance comparisons between traditional BSO and our proposed ASBSO at different dimensions show that ASBSO outperforms BSO entirely. The contrast between ASBSO and other well-known algorithms also indicates the superiority of ASBSO. (4) ASBSO is verified to be a competent and robust algorithm for different optimization problems.

The organization of this paper can be presented as follows. A brief introduction of BSO is given in Section \ref{section2}. Section \ref{section3} introduces the proposed ASBSO in details. The experimental results are shown in Section \ref{section4}. Some discussions are assigned in Section \ref{section5}. We conclude this paper in Section \ref{section6}.

\section{A Brief Introduction about BSO}
\label{section2}
BSO is a swarm intelligence algorithm which is inspired by the human brain storming behaviors and it assumes the individuals in search procedure as the ideas generated by the human brain. In its execution process, three main operations including the clustering, selection and generation of individuals are implemented to maintain the population diversity and convergence speed \cite{wang2017discovery}.

(A) Clustering: The original BSO uses a $k$-means clustering method to divide individuals in current population into several clusters according to the distance among individuals. They are continually updated, in the meantime, the distribution of individuals moves towards a smaller and smaller range  by the lapse of iterations via $k$-means method. Therefore, for a given problem, the clustering results can show the distribution of individuals in the search space.

(B) Selection: New individuals are generated based on one individual or the combination of two individuals. BSO controls the selection operation by presetting some parameters \cite{shi2011brain}. If a random value is smaller than a replacement parameter $p_c$ ($p_c=0.2$ in \cite{shi2011brain}), one cluster center is replaced by a randomly generated individual. Another parameter $p_g$ ($=0.8$) controls the number of selected individuals in the generation phase. If a random value is smaller than $p_g$, one cluster is selected, otherwise, two clusters are applied. After comparing with $p_g$, there are  two parameters $p_{c1}$ and $p_{c2}$ which further confirm the selected individuals from one and two clusters. To be specific, new generated individual from one cluster center or one general individual is decided by $p_{c1}$. Similarly, $p_{c2}$ determines new generated individual from two cluster centers or two general individuals.

(C) Generation: After implementing the selection of individuals, the generation method of BSO can be exhibited in Eqs. (\ref{bso}) and (\ref{xi}).
\begin{equation}
X_{new}=X+\xi \cdot N(0,1)
\label{bso}
\end{equation}
where $X$ and $X_{new}$ are the selected and newly generated individuals, respectively. Standard normal distribution $N(0,1)$ is used to generate a random variation. $\xi$ is a step length which is calculated in Eq. (\ref{xi}).

\begin{equation}
\xi=logsig(({{M}_{i}}/2-C_{i})/K)\cdot rand
\label{xi}
\end{equation}
where $logsig()$ means a logarithmic sigmoid  transfer function which ranges in the interval (0,1). ${M}_{i}$ and $C_{i}$ refer to the maximum iteration and current iteration. $K$ is used to change the scale of $logsig()$ function and $rand$ generates a random value in the interval $(0, 1)$. If the fitness value $f(X_{new})$ is better than $f(X)$, $X$ is replaced by $X_{new}$.

\section{ASBSO}
\label{section3}
\subsection{Motivation}
In the new individual generating operation of BSO introduced in  Section \ref{section2},  the search step length only varies with the current iteration number and lacks of flexibility,  thus it makes a poor search efficiency and robustness. BSO only applies an invariable scale parameter $K=20$ to render the search range to shrink during iterations, therefore the shrink is limited and inflexible. In ASBSO, an adaptive step length mechanism is motivated to alleviate this issue. Various optional scale parameters make BSO have adjustable search ranges instead of the traditional step length which only varies according to the current iteration number.  As ASBSO applies multiple step lengths in the search process, the probability of getting into the gorge or jumping out of the valley in the search landscape can be increased a lot.

As we described in Section \ref{section2} that BSO lacks of a powerful search ability and robustness, it is a motivation for us to alleviate these drawbacks. An example is shown below to make us further understand the utilization of search ability and robustness.

A popular approach to comprehensively observe the search ability and robustness of optimization algorithms in evolutionary community is to optimize benchmark functions. Some famous benchmark function suits such as 23 standard benchmark functions \cite{yao1999evolutionary}, CEC'05 \cite{suganthan2005problem}, CEC'13 \cite{ liang2013problem} and CEC'17 benchmark functions \cite{awad2016problem} have been widely used. These functions become more and more complicated and difficult in order to emulate the real world problems whose complexities increase in a geometric ratio. Therefore, the performances of optimization algorithms on benchmark functions have become an important standard to judge whether they can be implemented into practical applications or not. For instance, Fig. \ref{fig1} illustrates the 3D and contour graphs of F8 and F11 in CEC'13 function suit. F8 is a rotated Ackley's function which has the same properties of multi-modal, non-separable and asymmetrical as F11 does. In addition, the local optima's number of F11 is very huge. The global optimum of F8 seems to be in a gorge surrounded by many steep precipices. The entrance of this gorge is so narrow and secluded that it could easily be missed by a search step length which is beyond the distance between $X(t)$ and $X'(t)$. Once the entrance has been missed, individual could only find a mass of similar local optimum. It will take a lot of computational time to obtain another chance for exploiting the gorge where global optima hides. On the contrary, in F11, a step length smaller than the distance between $X(t)$ and $X'(t)$ means that the individual couldn't jump out of the valley of local optima and is hard to know the global optima lays just beside it. These are two representative cases which could happen not only in benchmark functions but also in real world. Therefore, it has become an urgent task to alleviate and solve them via proposing more suitable optimization algorithms .

To address the above issues, two main modifications including multiple step lengths and new memory mechanism are proposed in ASBSO. They are interpreted in the following subsections in detail.

\subsection{Multiple Step Lengths}
The parameter $K$ in Eq. (\ref{xi}) is used to change the scale of $logsig()$. In the strategy of multiple step lengths, different $K$ values listed in Table \ref{l} is applied to provide different scales to adjust the search step length. The strategies which have relatively small $K$ values indicate that they can provide a diffusion to search radius. It makes BSO be effective to explore the objective space and accelerate convergence. In the early search phase, optimization algorithm is required to have efficient exploration competence when facing the unknown search space. If we pay much attention to exploit local information before the whole space has been explored, the search cost will become very expensive and influence the solution quality \cite{molina2010memetic}. Thus, it's necessary to provide large search step length to effectively detect the region with promising solutions. While in exploitation phase, a local search which applies short step length is needed urgently to excavate solutions with a high accuracy. Therefore, strategies with relatively large $K$ values can improve solution quality in exploitation phase as large $K$ values generally lead to a localized search.

\begin{table}[!t]
\caption{Illustration of the flexible multiple search length strategy.}
\centering
\begin{tabular}{|c|c|c|c|c|c|}\hline
     &   Strategy 1 & Strategy 2 & Strategy 3 &...& Strategy M \\\hline
     $K$ &  $k$              &   $k+H$             &   $k+2H$   &...         &  $k+(M-1)H$  \\\hline
\end{tabular}
\label{l}
\end{table}

As we discussed that changeless $K$ value makes BSO only can shrink its search range according to the current iteration number while couldn't flexibly adjust step length to fit various search periods and problems, assigning multiple values to $K$ naturally equips BSO with flexible search ability to reply different situations.

\subsection{New Memory Mechanism}
To adaptively carry out multiple step lengths, we introduce an improved memory storing mechanism (IMS) which is originated from the success-failure-based memory structure (SFMS) \cite{song2017multiple,qin2009differential}, In SFMS, a success memory shown in Table \ref{successmemory} and a failure memory shown in Table \ref{failurememory} is applied to store the number of succeeding or failing to generate better solutions, respectively. In the beginning, $M$ strategies are randomly selected by roulette wheel selection method to generate new individuals. As Eqs. (\ref{ns}) and (\ref{nf}) shown, if the new individual $X'_{t-1}$ outperforms and replaces the old individual $X_{t-1}$, it is indicated as a success and let $\alpha_{j,t}$ equal to 1, where $j$ ($j=1,2,...,M$) refers to the used strategy and $t$ is the current iteration. If the opposite, it becomes a failure trial and $\beta_{j,t}$ equals to 1. If the iteration count is over the preset iteration length $L$ ($L=$50 is empirically set according to \cite{song2017multiple}), the first row of Tables \ref{successmemory} and  \ref{failurememory} will be removed to make space for the newest one. The selection of strategies is described as follows.

\begin{equation}
\alpha_{j,t}=\left \{ \begin{array}{cl}1,           & f(X'_{t-1})<f(X_{t-1})  \\0,    & otherwise \end{array}  \right.
 \label{ns}
 \end{equation}

 \begin{equation}
\beta_{j,t}=\left \{ \begin{array}{cl}0,           & f(X'_{t-1})<f(X_{t-1})  \\1,    & otherwise \end{array}  \right.
 \label{nf}
 \end{equation}

\begin{table}[!t]
\caption{Traditional Success Memory}
\centering
\begin{tabular}{|c|c|c|c|c|c|}
\hline Index  &\ Strategy 1   &\ Strategy 2   &\  Strategy 3  & ...  &\ Strategy M       \\
\hline   1    &\ $\alpha_{1,t-L}$   &\ $\alpha_{2,t-L}$    &\ $\alpha_{3,t-L}$ \ & ...  &   $\alpha_{M,t-L}$    \\
\hline  2   &\ $\alpha_{1,t-L+1}$ &\ $\alpha_{2,t-L+1}$  &\ $\alpha_{3,t-L+1}$  & ... & $\alpha_{M,t-L+1}$ \\
\hline  ...   & \    ...         &   \  ...         & \ ...   & ... & \    ...         \\
\hline  $L$   &\ $\alpha_{1,t-1}$     &\ $\alpha_{2,t-1}$     &\ $\alpha_{3,t-1}$   & ...   & $\alpha_{M,t-1}$    \\\hline
\end{tabular}
\label{successmemory}
\end{table}

\begin{table}[!t]
\centering
\caption{Traditional Failure Memory}
\begin{tabular}{|c|c|c|c|c|c|}\hline
Index  &\ Strategy 1   &\ Strategy 2    & \ Strategy 3  & ... &\ Strategy M       \\\hline
  1   &\ $\beta_{1,t-L}$\  &\ $\beta_{2,t-L}$\  &\ $\beta_{3,t-L}$\ & ...  & $\beta_{M,t-L}$  \\\hline
  2   &\ $\beta_{1,t-L+1}$ &\ $\beta_{2,t-L+1}$ & \ $\beta_{3,t-L+1}$  & ... & $\beta_{M,t-L+1}$  \\\hline
...   & \    ...        & \    ...        & \ ...   & ...  & \    ...        \\\hline
  $L$  &\ $\beta_{1,t-1}$  & \  $\beta_{2,t-1}$     & \  $\beta_{3,t-1}$    & ...  &   $\beta_{M,t-1}$    \\\hline
\end{tabular}
\label{failurememory}
\end{table}

The chosen probability of each strategy is calculated as shown in Eqs. (\ref{chooserate}) and (\ref{successrate}) after the memories record the results:

\begin{equation}
p_{j,t}=\frac{S_{j,t}}{\sum_{j=1}^{4}S_{j,t}}
\label{chooserate}
\end{equation}

\begin{equation}
S_{j,t}=\frac{\sum_{t-L}^{t-1}\alpha_{j,t}}{\sum_{t-L}^{t-1}\alpha_{j,t}+\sum_{t-L}^{t-1}\beta_{j,t}}+\delta
\label{successrate}
\end{equation}
where $p_{j,t}$ denotes the probability to use the $j$-th strategy in current iteration $t$ when $t>L$. $\sum_{t-L}^{t-1}\alpha_{j,t}$ calculates the total number of the $j$-th strategy successfully generating a new individual to replace $X_{t-1}$. $\sum_{t-L}^{t-1}\beta_{j,t}$ is the total number for the failure circumstances. Eq. (\ref{successrate}) calculates the success rate and $\delta=0.01$ is used for avoiding a null value. It is obvious that the strategy with higher success rate has a higher chance to be selected to generate new individuals.

However, the SFMS mechanism has one drawback that no matter how better a new individual obtained by a strategy, it only records $1$ in the success memory. One case is given to interpret this drawback in detail. Let's define that $D^1$ represents the improvement in fitness (if $f(X'_{t-1})<f(X_{t-1})$, $D=|f(X'_{t-1})-f(X_{t-1})|$) obtained by Strategy 1, $D^2$ is that obtained by Strategy 2, and so on. Supposing $D^1=2D^2$, which means Strategy 1 is suitable for the current search period and can find a much better solution than Strategy 2 does in one generation. However, they score the same points (both 1) in success memory which leads to same possibilities to be selected. This mechanism evidently has relatively low efficiency which causes a slowness in convergence speed, and further decrease the solution quality. To alleviate this issue, in IMS, the improvement value in fitness $D^j$ ($j$ indicates the executed strategy) is recorded into a success memory to replace the numbers of 0 and 1. In the meanwhile, failure memory is not implemented in the new mechanism, since we focus on the quality not quantity that each strategy obtains. If failure memory is applied, a poor search attempt may decrease the quality of solutions and hinder the evolutionary direction of algorithm. Table \ref{newsuccessmemory} shows the structure of IMS. Each improvement value $D_t^j$ in fitness obtained by strategy $j$ is stored in it. The selection possibility of strategy $j$ at iteration $t$ can be calculated by Eq. (\ref{su2}).

\begin{equation}
p^{new}_{j,t}=\frac{D^j_{t}}{\sum_{j=1}^{M}D^j_{t}}
\label{su2}
\end{equation}

\begin{table}[!t]
\caption{New Success Memory (IMS)}
\centering
\begin{tabular}{|c|c|c|c|c|c|}
\hline Index  &\ Strategy 1   &\ Strategy 2    & \ Strategy 3  & ...   &\ Strategy M       \\
\hline   1    &\ $D^1_{t-L}$   &\ $D^2_{t-L}$    &\ $D^3_{t-L}$  & ... &\   $D^M_{t-L}$    \\
\hline  2   &\ $D^1_{t-L+1}$ &\ $D^2_{t-L+1}$  &\ $D^3_{t-L+1}$   & ... &\ $D^M_{t-L+1}$ \\
\hline  ...   & \    ...        &   \  ...        & \ ...  & ... & \    ...        \\
\hline  $L$   &\ $D^1_{t-1}$     &\ $D^2_{t-1}$     &\ $D^3_{t-1}$    & ... & $D^M_{t-1}$   \\\hline
\end{tabular}
\label{newsuccessmemory}
\end{table}

Algorithm \ref{flowchart} illustrates the main procedures of ASBSO. In each generation of new individuals, a strategy $j$ is selected according to its selection possibility $p^{new}_{j,t}$ to produce a search step length. The new individual is generated by adding the step length to the selected $X$ by using Eq. (\ref{bso}) and its fitness is calculated. If the new individual is better than the old one, then it will replace the old one. In the meanwhile, the selected strategy is marked as a success trial. The improvement in fitness $D^j_{t}$ is stored in memory and the selection possibility for each strategy is updated.

\begin{algorithm}
\caption{Pseudo code of ASBSO.}
\label{flowchart}
Randomly generate a population with $N$ individuals\;
Calculate the fitness of each individual\;
\While{termination not satisfied
}{
Divide $N$ individuals into $C$ clusters by using $k-means$ clustering method\;
Choose the best individual in each cluster as the $center$\;
\If{$random(0,1)<p_c=0.2$
}{replace one cluster center by a randomly generated individual
}
\eIf{$random(0,1)<p_g=0.8$
}{select one cluster\;\eIf{$random(0,1)<p_{c1}=0.4$
}{choose the cluster $center$ as $X$
}{choose a randomly selected individual in the cluster as $X$}
}{randomly select two clusters\;\eIf{$random(0,1)<p_{c2}=0.5$}{choose the combination of two $centers$ as $X$
}{choose the combination of two randomly selected individuals in two clusters as $X$
}
}
Choose a strategy to generate a search step length according to Eq. (\ref{su2})\;
Generate new individual by adding the step length to the selected $X$ by using Eqs. (\ref{bso}) and (\ref{xi})\;
\If{new individual is better than old one
}{replace the old individual and update the memory
}
}
\end{algorithm}

\begin{table*}[!htbp]
\centering
\caption{Friedman test result for $H=10$, $20$ and $30$.}
\begin{tabular}{cccccc}\hline
Algorithm&Ranking &unadjusted $p$&$p_{Bonf}$&$p_{Holm}$&$p_{Hochberg}$ \\\hline
    $H=20$ vs.     &1.4298   \\\hline
$H=30$&2.2895&0.000004&0.000009&0.000009&0.000006\\\hline
$H=10$&2.2807&0.000006&0.000011&0.000009&0.000006\\\hline
\end{tabular}
\label{h}
\end{table*}

\begin{table*}[!htbp]
\centering
\caption{Experimental results of CEC'13 benchmark functions (F1-F28) using BSO and ASBSO at $D=10$ and $D=30$.}
\begin{tabular}{c|c|c|c|c|c}\hline
     \multicolumn{3}{c}{$D$=10}                                                                            &                 \multicolumn{3}{c}{$D$=30}      \\\hline
  &BSO                                           & ASBSO                                &   & BSO                       & ASBSO \\\hline
  & Mean (Std Dev)                        & Mean (Std Dev)                   &     & Mean (Std Dev)             & Mean (Std Dev)  \\\hline
F1	&                 -1.40E+03 (0.00E+00)	     &	-1.40E+03 (0.00E+00)             &F1      	&                  -1.40E+03 (4.22E-14)	     &	-1.40E+03 (1.98E-13) \\\hline
F2	&	           6.79E+04 (5.55E+04)	&	\bf{3.85E+04 (3.29E+04)}   &F2  &                    1.54E+06 (4.79E+05)	     &	1.54E+06 (4.26E+05)\\\hline
F3	&		4.01E+07 (7.33E+07)	&	\bf{2.92E+07 (4.94E+07)}         &F3 &                1.11E+08 (1.74E+08)		     &	\bf{8.47E+07 (8.64E+07)}\\\hline
F4	&			7.58E+03 (4.24E+03)	&	\bf{6.00E+03 (3.58E+03)}	 &F4&                 2.17E+04 (5.65E+03)	     &	\bf{5.08E+03 (2.25E+03)}\\\hline
F5	&			-1.00E+03 (1.35E-04)	&	-1.00E+03 (1.88E-04)	&F5&          -1.00E+03 (1.52E-03)     &	-1.00E+03 (2.98E-03)	\\\hline
F6	&			\bf{-8.97E+02 (2.65E+00)}&	 -8.93E+02 (4.23E+00)&F6	&                 \bf{ -8.66E+02 (2.47E+01)}	     &	-8.64E+02 (2.76E+01)\\\hline
F7	&			-7.05E+02 (3.36E+01)&	\bf{-7.24E+02 (3.17E+01)}	&F7&                -6.71E+02 (7.57E+01)	     &	\bf{-7.08E+02 (3.90E+01)}\\\hline
F8	&			-6.80E+02(9.14E-02)	&	-6.80E+02 (9.48E-02)		&F8     &	-6.79E+02 (7.60E-02)	&	-6.79E+02 (6.70E-02) \\\hline
F9	&		       -5.93E+02 (1.41E+00)	&	\bf{-5.94E+02 (1.48E+00)}		&F9     &	-5.68E+02 (2.90E+00)&	\bf{-5.71E+02 (2.58E+00)} \\\hline
F10	&			-5.00E+02 (3.15E-02)	&	-5.00E+02 (4.81E-02)		&F10     &	-5.00E+02 (1.93E-01)&	-5.00E+02 (5.35E-02)  \\\hline
F11	&			-3.42E+02 (1.90E+01)	&	\bf{-3.53E+02 (2.38E+01)}		&F11     &	6.40E+01 (7.12E+01)&	\bf{-1.82E+02 (5.40E+01)}  \\\hline
F12	&			-2.46E+02(1.86E+01)	&	-2.46E+02 (2.17E+01)		   &F12  &	2.06E+02 (8.43E+01)&	\bf{-7.64E+01 (4.85E+01) } \\\hline
F13	&			-1.30E+02 (2.14E+01)	&	\bf{-1.31E+02 (2.09E+01)}	&          F13    &	3.55E+02 (8.77E+01)&	\bf{1.30E+02 (6.54E+01)}  \\\hline
F14	&			1.04E+03 (2.33E+02)	&	\bf{8.73E+02 (2.96E+02)}		&F14     &	3.88E+03 (5.17E+02)&	\bf{3.68E+03 (4.56E+02} \\\hline
F15	&			1.17E+03 (2.79E+02)	&	\bf{1.05E+03 (2.78E+02)}		  &F15   &	4.25E+03 (5.57E+02)&	\bf{3.88E+03 (5.74E+02)}  \\\hline
F16	&			2.00E+02 (2.02E-02)	&	2.00E+02 (7.00E-02)		 &F16    &	2.00E+02 (4.14E-02)&	2.00E+02 (1.13E-01)  \\\hline
F17	&			3.55E+02 (1.86E+01)	&	\bf{3.39E+02 (1.13E+01)}		 &F17    &	7.31E+02 (8.13E+01)&	\bf{5.28E+02 (5.13E+01)}  \\\hline
F18	&			4.49E+02 (2.27E+01)	&	\bf{4.39E+02 (1.28E+01)}		&F18     &	7.41E+02 (5.30E+01)&	\bf{5.98E+02 (2.85E+01)} \\\hline
F19	&			5.02E+02 (5.95E-01)	&	\bf{5.01E+02 (4.27E-01)}	 &F19    &	5.09E+02 (2.04E+00)&	\bf{5.04E+02 (7.60E-01)}  \\\hline
F20	&			6.04E+02 (6.46E-01)	&	\bf{6.03E+02 (5.66E-01)}	   &F20  &	6.14E+02 (1.77E-01)&	6.14E+02 (2.94E-01)  \\\hline
F21	&			1.10E+03 (4.63E-13)	&	1.10E+03 (2.16E-11)	 &F21    &	1.03E+03 (8.91E+01)&	\bf{1.02E+03 (8.42E+01)}  \\\hline
F22	&			2.24E+03 (3.35E+02)	&	\bf{2.05E+03 (2.72E+02)}		&F22     &	6.04E+03 (7.04E+02)&	\bf{5.36E+03 (4.70E+02)}  \\\hline
F23	&			\bf{2.19E+03 (3.06E+02)}	&	2.28E+03 (3.23E+02)		 &F23    &	\bf{5.98E+03 (7.32E+02)}&	6.00E+03 (7.84E+02)  \\\hline
F24	&			1.22E+03 (1.16E+01)	&	1.22E+03 (1.37E+01)		 &F24    &	1.33E+03 (2.25E+01)&	\bf{1.31E+03 (2.60E+01)}  \\\hline
F25	&			1.32E+03 (4.24E+00)	&	1.32E+03 (1.99E+01)		 &F25    &	1.46E+03 (2.50E+01)&	\bf{1.41E+03 (1.03E+01)}  \\\hline
F26	&			1.39E+03 (3.26E+01)	&	1.39E+03 (2.64E+01)		&F26     &	1.50E+03 (8.60E+01)&	\bf{1.46E+03 (7.91E+01)}  \\\hline
F27	&			1.81E+03 (1.12E+02)	&	\bf{1.77E+03 (1.18E+02)}		 &F27    &	2.49E+03 (9.32E+01)&	\bf{2.42E+03 (1.07E+02)}  \\\hline
F28	&			2.26E+03 (7.51E+01)	&	\bf{2.17E+03 (1.78E+02)}		 &F28    &	5.73E+03 (5.38E+02)&	\bf{2.03E+03 (8.02E+02)}  \\\hline
\end{tabular}
\label{D10D30}
\end{table*}

\begin{table*}[!htbp]
\centering
\caption{Experimental results of CEC'13 benchmark functions (F1-F28) using BSO and ASBSO at $D=50$ and $D=100$.}
\begin{tabular}{c|c|c|c|c|c}\hline
     \multicolumn{3}{c}{$D$=50}                                                                            &                 \multicolumn{3}{c}{$D$=100}      \\\hline
  &BSO                                           & ASBSO                                &   & BSO                       & ASBSO \\\hline
  & Mean (Std Dev)                        & Mean (Std Dev)                   &     & Mean (Std Dev)             & Mean (Std Dev)  \\\hline
F1	&               -1.40E+03 (1.25E-06)	     &	-1.40E+03 (1.54E-02)             &F1      	&                  -1.40E+03 (5.99E-02)	     &	-1.40E+03 (8.62E-01) \\\hline
F2	&	           \bf{2.25E+06 (7.00E+05)}	     &	2.33E+06 (7.15E+05)  &F2  &                     \bf{1.03E+07 (2.17E+06)	}     &	1.25E+07 (2.24E+06)\\\hline
F3	&		 \bf{2.09E+08 (1.52E+08)}	     &	2.68E+08 (1.55E+08)      &F3 &                  \bf{1.45E+09 (5.98E+08)	}     &	2.73E+09 (1.39E+09)\\\hline
F4	&			  1.58E+04 (5.04E+03)	     &	\bf{7.93E+03 (2.63E+03)} &F4&                  6.38E+03 (1.79E+03)	     &	\bf{2.12E+03 (7.60E+02)}\\\hline
F5	&			  -1.00E+03 (4.09E-03)	     &	-1.00E+03 (1.31E-02)&F5&           -1.00E+03 (2.86E-02)	     &	-1.00E+03 (5.38E-01)	\\\hline
F6	&			 \bf{ -8.27E+02 (3.53E+01)}	     &	-8.17E+02 (3.26E+01)&F6	&                   \bf{-7.05E+02 (4.96E+01)}	     &	-6.73E+02 (5.28E+01)\\\hline
F7	&			  -6.37E+02 (6.43E+01)	     &	\bf{-6.82E+02 (3.23E+01)}	&F7&                 -6.71E+02 (2.76E+01)	     &	\bf{-7.05E+02 (2.05E+01)}\\\hline
F8	&			  -6.79E+02 (5.55E-02)	     &	-6.79E+02 (4.13E-02)	&F8     &	  -6.79E+02 (4.28E-02)	     &	-6.79E+02 (4.98E-02)\\\hline
F9	&		        -5.43E+02 (3.70E+00)	     &	\bf{-5.49E+02 (4.38E+00)}	&F9     &	  -4.72E+02 (4.81E+00)	     &	\bf{-4.82E+02 (6.31E+00)} \\\hline
F10	&			  -4.99E+02 (1.71E-01)	     &	-4.99E+02 (4.52E-01)	&F10     &	  \bf{-4.95E+02 (6.45E-01)	}     &	-4.94E+02 (2.06E+00) \\\hline
F11	&			  3.27E+02 (9.56E+01)	     &	\bf{2.22E+02 (6.95E+01)}		&F11     &	  1.54E+03 (1.94E+02)	     &	\bf{1.40E+03 (1.72E+02)}  \\\hline
F12	&			  \bf{4.63E+02 (1.17E+02)}	     &	4.66E+02 (1.08E+02)		   &F12  &	  1.89E+03 (2.56E+02)	     &	\bf{1.83E+03 (2.34E+02)} \\\hline
F13	&			  \bf{6.84E+02 (1.04E+02)}	     &	6.98E+02 (1.22E+02) &F13&                  2.29E+03 (2.18E+02)	     &	\bf{2.15E+03 (2.15E+02)}\\\hline
F14	&			 7.00E+03 (7.93E+02)	     &	\bf{6.70E+03 (7.94E+02)} &F14&                  1.52E+04 (1.12E+03)	     &	\bf{1.46E+04 (9.60E+02)}\\\hline
F15	&			   7.93E+03 (1.01E+03)	     &	\bf{7.41E+03 (5.66E+02)} &F15&                  1.53E+04 (1.27E+03)	     &	\bf{1.45E+04 (1.11E+03)}\\\hline
F16	&			 2.00E+02 (9.01E-02)	     &	2.00E+02 (1.55E-01) &F16&                  2.01E+02 (1.39E-01)	     &	\bf{2.00E+02 (3.84E-01)}\\\hline
F17	&			  1.14E+03 (9.10E+01)	     &	\bf{7.69E+02 (6.23E+01)} &F17&                  2.36E+03 (1.89E+02)	     &	\bf{1.36E+03 (1.58E+02)}\\\hline
F18	&			 1.01E+03 (7.72E+01)	     &	\bf{7.38E+02 (5.20E+01)} &F18&                  1.94E+03 (1.47E+02)	     &	\bf{1.25E+03 (1.62E+02)}\\\hline
F19	&			 5.16E+02 (2.43E+00)	     &	\bf{5.11E+02 (2.81E+00)} &F19&                  5.45E+02 (4.78E+00)	     &	\bf{5.33E+02 (9.08E+00)}\\\hline
F20	&			 6.24E+02 (4.50E-01)	     &	6.24E+02 (4.15E-01) &F20&                  6.50E+02 (2.11E-14)	     &	6.50E+02 (4.71E-12)\\\hline
F21	&			 1.65E+03 (3.18E+02)	     &	\bf{1.44E+03 (4.27E+02)} &F21&                  1.14E+03 (6.13E+01)	     &	1.14E+03 (6.00E+01)\\\hline
F22	&			 1.08E+04 (1.36E+03)	     &	\bf{1.04E+04 (1.20E+03)} &F22&                  2.34E+04 (1.87E+03)	     &	\bf{2.22E+04 (2.42E+03)}\\\hline
F23	&			 1.09E+04 (1.07E+03)	     &	\bf{1.04E+04 (1.35E+03)} &F23&                  2.21E+04 (1.59E+03)	     &	\bf{2.19E+04 (1.57E+03)}\\\hline
F24	&			 1.44E+03 (6.62E+01)	     &	\bf{1.38E+03 (2.00E+01)} &F24&                  2.44E+03 (5.16E+02)	     &	\bf{1.72E+03 (2.97E+02)}\\\hline
F25	&			 1.59E+03 (3.53E+01)	     &	\bf{1.58E+03 (3.03E+01)} &F25&                  1.96E+03 (1.01E+02)	     &	1.96E+03 (1.12E+02)\\\hline
F26	&			 1.64E+03 (6.94E+01)	     &	\bf{1.60E+03 (9.24E+01)} &F26&                  1.85E+03 (1.88E+01)	     &	\bf{1.82E+03 (2.40E+01)}\\\hline
F27	&			 3.48E+03 (1.65E+02)	     &	\bf{3.26E+03 (1.50E+02)} &F27&                  5.57E+03 (2.44E+02)	     &	\bf{5.10E+03 (2.59E+02)}\\\hline
F28	&			 9.11E+03 (7.24E+02)	     &	\bf{8.92E+03 (5.80E+02)} &F28&                  1.97E+04 (1.72E+03)	     &	\bf{1.93E+04 (1.48E+03)}\\\hline
\end{tabular}
\label{D50D100}
\end{table*}

\begin{table}[!htbp]
\centering\small
\caption{Results obtained by the Wilcoxon signed-rank test for ASBSO vs. BSO on CEC'13.}
\begin{tabular}{
c|c|c|c|c|c}
\hline
Dimension & $R^{+}$ & $R^{-}$ & $p$-value &$\alpha$=0.05 & $\alpha$=0.01 \\ \hline
10 & 330.5 & 75.5 & 2.782E-3 & YES& YES\\ \hline
30 & 319.0 & 59.0 & 1.132E-3 & YES& YES\\ \hline
50 & 319.0 & 87.0 & 7.072E-3 &  YES& YES\\ \hline
100 & 321.0 & 85.0 & 6.06E-3 &  YES& YES\\ \hline
\end{tabular}
\label{wilcoxon13}
\end{table}

\begin{table*}[!htbp]
\centering
\caption{Experimental results of CEC'17 benchmark functions (F29-F57) using BSO and ASBSO at $D=10$ and $D=30$.}
\begin{tabular}{c|c|c|c|c|c}\hline
     \multicolumn{3}{c}{$D$=10}                                                                            &                 \multicolumn{3}{c}{$D$=30}      \\\hline
  &BSO                                           & ASBSO                                &   & BSO                       & ASBSO \\\hline
  & Mean (Std Dev)                        & Mean (Std Dev)                   &     & Mean (Std Dev)             & Mean (Std Dev)  \\\hline
F29	&	8.95E+02	(	1.09E+03	)	&	\bf{8.60E+02	(	1.12E+03	)}	&	F29	&	2.47E+03	(	1.95E+03	)	&	\bf{2.21E+03	(2.00E+03)}\\\hline
F30	&	3.00E+02	(	0.00E+00	)	&	3.00E+02	(	1.67E-09	)	&	F30	&	5.34E+02	(	2.66E+02	)	&	\bf{3.95E+02	(1.10E+02)}\\\hline
F31	&	4.04E+02	(	7.05E+00	)	&	\bf{4.03E+02	(	1.50E+00	)}	&	F31	&	\bf{4.67E+02	(	2.19E+01	)}	&	4.72E+02	(2.92E+01)\\\hline
F32	&	5.35E+02	(	1.43E+01	)	&	\bf{5.34E+02	(	1.30E+01	)}	&	F32	&	6.87E+02	(	4.05E+01	)	&\bf{6.86E+02	(3.45E+01)}\\\hline
F33 &	6.24E+02	(	8.60E+00	)	&	6.24E+02	(	7.25E+00	)	&	F33	&	6.52E+02	(	7.01E+00	)	&	\bf{6.51E+02	(7.77E+00)}\\\hline
F34	&	7.57E+02	(	1.79E+01	)	&	\bf{7.54E+02	(	2.15E+01	)}	&	F34	&	\bf{1.15E+03	(	9.65E+01	)}	&	1.16E+03	(9.94E+01)\\\hline
F35	&	8.23E+02	(	9.45E+00	)	&	\bf{8.22E+02	(	9.17E+00	)}	&	F35	&	9.47E+02	(	2.82E+01	)	&	\bf{9.41E+02	(3.19E+01)}\\\hline
F36	&	1.09E+03	(	1.58E+02	)	&	1.09E+03	(	1.03E+02	)	&	F36	&	3.98E+03	(	6.95E+02	)	&	\bf{3.93E+03	(6.39E+02)}\\\hline
F37	&	2.10E+03	(	2.57E+02	)	&	\bf{2.08E+03	(	3.22E+02	)}	&	F37	&	5.30E+03	(	5.16E+02	)	&	\bf{5.20E+03	(5.67E+02)}\\\hline
F38	&	\bf{1.15E+03	(	3.21E+01	)}	&	1.16E+03	(	3.52E+01	)	&	F38	&	1.23E+03	(	4.05E+01	)	&	1.23E+03	(4.75E+01)\\\hline
F39	&	1.19E+05	(	1.29E+05	)	&	\bf{5.90E+04	(	5.40E+04	)}	&	F39	&	1.77E+06	(	1.25E+06	)	&	\bf{1.41E+06	(8.00E+05)}\\\hline
F40	&	8.96E+03	(	5.30E+03	)	&	\bf{7.80E+03	(	5.55E+03	)}	&	F40	&	5.36E+04	(	2.85E+04	)	&	\bf{5.04E+04	(2.64E+04)}\\\hline
F41	&	1.72E+03	(	1.06E+03	)	&	\bf{1.71E+03	(	3.43E+02	)}	&	F41	&	\bf{6.40E+03	(	4.66E+03	)}	&	7.08E+03	(5.23E+03)\\\hline
F42	&	4.12E+03	(	1.91E+03	)	&	\bf{4.10E+03	(	3.34E+03	)}	&	F42	&	\bf{2.95E+04	(	1.61E+04	)}	&	3.01E+04	(2.25E+04)\\\hline
F43	&	1.92E+03	(	1.12E+02	)	&	\bf{1.87E+03	(	1.23E+02	)}	&	F43	&	3.20E+03	(	4.24E+02	)	&	\bf{3.01E+03	(2.25E+02)}\\\hline
F44	&	1.77E+03	(	4.19E+01	)	&	1.77E+03	(	4.48E+01	)	&	F44	&	2.48E+03	(	2.56E+02	)	&	\bf{2.40E+03	(2.44E+02)}\\\hline
F45	&	1.03E+04	(	1.26E+04	)	&	\bf{9.78E+03	(	1.01E+04	)}	&	F45	&	\bf{1.21E+05	(	1.03E+05	)}	&	1.23E+05	(1.21E+05)\\\hline
F46	&	3.28E+03	(	2.12E+03	)	&	\bf{3.15E+03	(	1.71E+03	)}	&	F46	&	1.52E+05	(	6.43E+04	)	&	\bf{1.25E+05	(6.31E+04)}\\\hline
F47	&	2.13E+03	(	6.33E+01	)	&	2.13E+03	(	6.22E+01	)	&	F47	&	2.67E+03	(	1.74E+02	)	&	2.67E+03	(2.17E+02)\\\hline
F48	&	2.29E+03	(	6.44E+01	)	&	\bf{2.27E+03	(	5.96E+01	)}	&	F48	&	2.50E+03	(	4.48E+01	)	&	\bf{2.49E+03	(3.14E+01)}\\\hline
F49	&	2.30E+03	(	1.03E+01	)	&	2.30E+03	(	1.14E+01	)	&	F49	&	6.03E+03	(	1.77E+03	)	&	\bf{5.79E+03	(2.04E+03)}\\\hline
F50	&	2.70E+03	(	3.40E+01	)	&	\bf{2.69E+03	(	2.86E+01	)}	&	F50	&	3.29E+03	(	1.27E+02	)	&	\bf{3.26E+03	(1.24E+02)}\\\hline
F51	&	2.78E+03	(	1.19E+02	)	&	\bf{2.73E+03	(	1.46E+02	)}	&	F51	&	3.50E+03	(	1.13E+02	)	&	\bf{3.49E+03	(9.56E+01)}\\\hline
F52	&	\bf{2.92E+03	(	2.25E+01	)}	&	2.93E+03	(	2.19E+01	)	&	F52	&	2.89E+03	(	1.46E+01	)	&	2.89E+03	(1.25E+01)\\\hline
F53	&	\bf{3.33E+03	(	3.90E+02	)}	&	3.34E+03	(	3.37E+02	)	&	F53	&	8.16E+03	(	1.57E+03	)	&	\bf{7.84E+03	(1.80E+03)}\\\hline
F54	&	\bf{3.16E+03	(	3.16E+01	)}	&	3.17E+03	(	3.40E+01	)	&	F54	&	\bf{3.82E+03	(	2.91E+02	)}	&	3.85E+03	(2.17E+02)\\\hline
F55	&	\bf{3.21E+03	(	8.38E+01	)}	&	3.23E+03	(	1.81E+02	)	&	F55	&	3.21E+03	(	2.57E+01	)	&	\bf{3.18E+03	(3.60E+01)}\\\hline
F56	&	3.26E+03	(	8.93E+01	)	&	3.26E+03	(	6.44E+01	)	&	F56	&	\bf{4.38E+03	(	2.79E+02	)}	&	4.40E+03	(3.39E+02)\\\hline
F57	&	\bf{5.81E+04	(	3.46E+04	)}	&	2.95E+05	(	6.09E+05	)	&	F57	&	5.74E+05	(	3.50E+05	)	&	\bf{5.16E+05	(2.90E+05)} \\\hline
\end{tabular}
\label{17D10D30}
\end{table*}

\begin{table*}[!htbp]
\centering
\caption{Experimental results of CEC'17 benchmark functions (F29-F57) using BSO and ASBSO at $D=50$ and $D=100$.}
\begin{tabular}{c|c|c|c|c|c}\hline
     \multicolumn{3}{c}{$D$=50}                                                                            &                 \multicolumn{3}{c}{$D$=100}      \\\hline
  &BSO                                           & ASBSO                                &   & BSO                       & ASBSO \\\hline
  & Mean (Std Dev)                        & Mean (Std Dev)                   &     & Mean (Std Dev)             & Mean (Std Dev)  \\\hline
F29	&	2.30E+03	(	2.15E+03	)	&	\bf{1.20E+03	(	1.21E+03	)}	&	F29	&	5.23E+05	(	1.62E+05	)	&	\bf{3.34E+05	(	9.55E+05	)}\\\hline
F30	&	7.44E+03	(	2.73E+03	)	&	\bf{6.53E+03	(	2.25E+03	)}	&	F30	&	8.99E+04	(	1.69E+04	)	&	\bf{8.88E+04	(	1.71E+04	)}\\\hline
F31	&	5.47E+02	(	5.86E+01	)	&	\bf{5.42E+02	(	5.06E+01	)}	&	F31	&	6.81E+02	(	4.83E+01	)	&	\bf{6.80E+02	(	4.62E+01	)}\\\hline
F32	&	8.17E+02	(	3.72E+01	)	&	\bf{8.12E+02	(	4.79E+01	)}	&	F32	&	1.31E+03	(	7.77E+01	)	&	\bf{1.30E+03	(	8.56E+01	)}\\\hline
F33	&	6.61E+02	(	4.98E+00	)	&	\bf{6.60E+02	(	5.32E+00	)}	&	F33	&	6.65E+02	(	4.33E+00	)	&	\bf{6.64E+02	(	3.50E+00	)}\\\hline
F34	&	1.66E+03	(	1.22E+02	)	&	\bf{1.65E+03	(	1.34E+02	)}	&	F34	&	3.34E+03	(	2.43E+02	)	&	\bf{3.32E+03	(	2.94E+02	)}\\\hline
F35	&	1.13E+03	(	4.31E+01	)	&	1.13E+03	(	4.61E+01	)	&	F35	&	1.73E+03	(	8.70E+01	)	&	\bf{1.69E+03	(	7.42E+01	)}\\\hline
F36	&	1.12E+04	(	1.32E+03	)	&\bf{1.09E+04	(	1.56E+03	)}	&	F36	&	2.69E+04	(	3.07E+03	)	&	\bf{2.43E+04	(	3.87E+03	)}\\\hline
F37	&	\bf{8.17E+03	(	1.01E+03	)}	&	8.24E+03	(	8.64E+02	)	&	F37	&	1.65E+04	(	1.16E+03	)	&	\bf{1.62E+04	(	1.14E+03	)}\\\hline
F38	&	1.31E+03	(	3.87E+01	)	&	\bf{1.30E+03	(	4.78E+01	)}	&	F38	&	2.45E+03	(	2.72E+02	)	&	\bf{2.39E+03	(	1.41E+02	)}\\\hline
F39	&	1.15E+07	(	6.98E+06	)	&	1.15E+07	(	5.00E+06	)	&	F39	&	7.74E+07	(	1.56E+07	)	&	\bf{6.92E+07	(	1.34E+07	)}\\\hline
F40	&	5.26E+04	(	2.34E+04	)	&	\bf{5.86E+04	(	2.70E+04	)}	&	F40	&	3.84E+04	(	1.60E+04	)	&	\bf{3.83E+04	(	1.16E+04	)}\\\hline
F41	&	4.11E+04	(	2.81E+04	)	&	\bf{3.17E+04	(	1.70E+04	)}	&	F41	&	3.54E+05	(	1.24E+05	)	&	\bf{2.81E+05	(	1.20E+05	)}\\\hline
F42	&	3.06E+04	(	2.08E+04	)	&	\bf{2.50E+04	(	1.47E+04	)}	&	F42	&	\bf{3.07E+04	(	1.29E+04	)}	&	3.11E+04	(	1.08E+04	)\\\hline
F43	&	3.86E+03	(	4.72E+02	)	&	\bf{3.82E+03	(	4.50E+02	)}	&	F43	&	6.76E+03	(	7.95E+02	)	&	\bf{6.73E+03	(	7.46E+02	)}\\\hline
F44	&	3.61E+03	(	3.70E+02	)	&	3.61E+03	(	3.71E+02	)	&	F44	&	5.58E+03	(	6.39E+02	)	&	\bf{5.49E+03	(	5.70E+02	)}\\\hline
F45	&	3.53E+05	(	1.31E+05	)	&	\bf{2.97E+05	(	9.96E+04	)}	&	F45	&	5.04E+05	(	1.67E+05	)	&	\bf{5.03E+05	(	2.06E+05	)}\\\hline
F46	&	5.35E+05	(	2.17E+05	)	&	\bf{4.09E+05	(	1.92E+05	)}	&	F46	&	2.41E+06	(	1.20E+06	)	&	\bf{2.24E+06	(	1.07E+06	)}\\\hline
F47	&	3.63E+03	(	2.15E+02	)	&	\bf{3.42E+03	(	3.69E+02	)}	&	F47	&	\bf{5.69E+03	(	4.49E+02	)}	&	5.76E+03	(	5.01E+02	)\\\hline
F48	&	2.75E+03	(	6.79E+01	)	&	\bf{2.73E+03	(	7.92E+01	)}	&	F48	&	\bf{3.99E+03	(	1.67E+02	)}	&	4.02E+03	(	1.78E+02	)\\\hline
F49	&	1.03E+04	(	6.94E+02	)	&	\bf{9.86E+03	(	6.91E+02	)}	&	F49	&	\bf{1.88E+04	(	9.48E+02	)}	&	1.91E+04	(	1.16E+03	)\\\hline
F50	&	\bf{4.00E+03	(	1.52E+02	)}	&	4.01E+03	(	2.17E+02	)	&	F50	&	\bf{5.44E+03	(	3.08E+02	)}	&	5.50E+03	(	2.81E+02	)\\\hline
F51	&	\bf{4.14E+03	(	1.37E+02	)}	&	4.16E+03	(	2.01E+02	)	&	F51	&	6.35E+03	(	5.21E+02	)	&	\bf{6.22E+03	(	4.96E+02	)}\\\hline
F52	&	\bf{2.95E+03	(	2.86E+01	)}	&	3.07E+03	(	2.77E+01	)	&	F52	&	3.27E+03	(	7.41E+01	)	&	\bf{3.32E+03	(	4.75E+01	)}\\\hline
F53	&	1.29E+04	(	2.04E+03	)	&	\bf{1.26E+04	(	2.06E+03	)}	&	F53	&	\bf{3.07E+04	(	1.45E+03	)}	&	3.08E+04	(	1.90E+03	)\\\hline
F54	&	5.50E+03	(	5.93E+02	)	&	\bf{5.43E+03	(	4.31E+02	)}	&	F54	&	7.79E+03	(	1.49E+03	)	&	\bf{7.46E+03	(	1.28E+03	)}\\\hline
F55	&	\bf{3.29E+03	(	2.16E+01	)}	&	3.31E+03	(	2.64E+01	)	&	F55	&	\bf{3.36E+03	(	3.28E+01	)}	&	3.38E+03	(	2.36E+01	)\\\hline
F56	&	5.61E+03	(	3.71E+02	)	&	\bf{5.59E+03	(	4.17E+02	)}	&	F56	&	9.11E+03	(	5.95E+02	)	&	\bf{9.01E+03	(	6.78E+02	)}\\\hline
F57	&	\bf{1.67E+07	(	1.54E+06	)}	&	1.74E+07	(	1.97E+06	)	&	F57	&	1.26E+07	(	4.04E+06	)	&	\bf{1.03E+07	(	3.48E+06	)}\\\hline
\end{tabular}
\label{17D50D100}
\end{table*}

\begin{table}[!t]
\centering\small
\caption{Results obtained by the Wilcoxon signed-rank test for ASBSO vs. BSO on CEC'17.}
\begin{tabular}{
c|c|c|c|c|c}\hline
Dimension & $R^{+}$ & $R^{-}$ & $p$-value &$\alpha$=0.05 & $\alpha$=0.01 \\ \hline
10 & 295.0 & 111.0 &  3.576E-2 &  YES& NO\\ \hline
30 & 300.5 & 105.5 & 2.555E-2 & YES& NO\\ \hline
50 & 302.0 & 104.0 & 2.322E-2& YES& NO \\ \hline
100 & 338.0 & 97.0 & 8.008E-3 &  YES& YES\\ \hline
\end{tabular}
\label{wilcoxon17}
\end{table}

\begin{table*}[p]
\caption{Experimental results of CEC'13 (F1-F28) using ASBSO, CGSA-M, MABC, ABC, DE, WOA and SCA.}
\centering
\scalebox{1.0}{
\begin{tabular}{lcccc}\\\hline
Algorithm &  F1  &  F2  &  F3  &  F4   \\\hline
ASBSO & -1.40E+03 $\pm$ 1.98E-13 & \bf{1.54E+06 $\pm$ 4.26E+05} & \bf{8.47E+07 $\pm$ 8.64E+07} & \bf{5.08E+03 $\pm$ 2.25E+03}  \\\hline
CGSA-M  & -1.40E+03 $\pm$ 0.00E+00 & 7.31E+06 $\pm$ 1.14E+06 & 5.82E+09 $\pm$ 1.89E+09 & 6.60E+04 $\pm$ 4.02E+03 \\\hline
MABC  & -1.40E+03 $\pm$ 0.00E+00 & 2.06E+08 $\pm$ 3.74E+07 & 6.86E+10 $\pm$ 1.90E+10 & 7.36E+04 $\pm$ 8.16E+03  \\\hline
ABC  & -1.40E+03 $\pm$ 1.03E-13 & 2.19E+08 $\pm$ 3.21E+07 & 6.66E+10 $\pm$ 1.59E+10 & 7.03E+04 $\pm$ 9.91E+03  \\\hline
DE & -7.26E+02 $\pm$ 3.32E+02 & 1.13E+08 $\pm$ 2.05E+07 & 1.32E+10 $\pm$ 2.63E+09 & 5.72E+04 $\pm$ 9.55E+03  \\\hline
WOA  & -1.40E+03 $\pm$ 1.60E-01 & 3.47E+07 $\pm$ 1.64E+07 & 1.35E+10 $\pm$ 7.98E+09 & 5.69E+04 $\pm$ 2.12E+04  \\\hline
SCA  & 9.08E+03 $\pm$ 1.72E+03 & 1.31E+08 $\pm$ 3.26E+07 & 3.12E+10 $\pm$ 8.79E+09 & 3.17E+04 $\pm$ 6.61E+03  \\\hline
  Algorithm &  F5  &  F6  &  F7  &  F8   \\\hline
ASBSO & -1.00E+03 $\pm$ 2.98E-03 & -8.64E+02 $\pm$ 2.46E+01 & -7.08E+02 $\pm$ 3.90E+01 & -6.79E+02 $\pm$ 6.70E-02  \\\hline
CGSA-M  & -1.00E+03 $\pm$ 8.64E-13 & -8.35E+02 $\pm$ 1.49E+01 &\bf{ -7.28E+02 $\pm$ 3.33E+01} & -6.79E+02 $\pm$ 5.31E-02  \\\hline
MABC  & -1.50E+02 $\pm$ 1.44E+02 & \bf{-8.67E+02 $\pm$ 1.32E+01} & -5.73E+02 $\pm$ 2.37E+01 & -6.79E+02 $\pm$ 7.13E-02  \\\hline
ABC  & -4.83E+01 $\pm$ 1.43E+02 & -8.57E+02 $\pm$ 1.25E+01 & -5.76E+02 $\pm$ 2.82E+01 & -6.79E+02 $\pm$ 4.36E-02  \\\hline
DE & -9.32E+02 $\pm$ 1.42E+01 & -7.37E+02 $\pm$ 2.44E+01 & -6.88E+02 $\pm$ 1.16E+01 & -6.79E+02 $\pm$ 4.22E-02  \\\hline
WOA  & -9.20E+02 $\pm$ 1.77E+01 & -7.99E+02 $\pm$ 3.72E+01 & -9.41E+01 $\pm$ 2.06E+03 & -6.79E+02 $\pm$ 4.45E-02  \\\hline
SCA  & 9.15E+02 $\pm$ 3.40E+02 & -2.06E+02 $\pm$ 2.22E+02 & -6.27E+02 $\pm$ 4.28E+01 & -6.79E+02 $\pm$ 6.99E-02  \\\hline
  Algorithm &  F9  &  F10  &  F11  &  F12   \\\hline
ASBSO & \bf{-5.71E+02 $\pm$ 3.23E+00} & \bf{-5.00E+02 $\pm$ 5.35E-02} & -1.82E+02 $\pm$ 5.40E+01 & \bf{-7.64E+01 $\pm$ 4.85E+01} \\\hline
 CGSA-M & -5.69E+02 $\pm$ 3.41E+00 & -5.00E+02 $\pm$ 7.86E-02 & -1.11E+02 $\pm$ 2.41E+01 & 3.68E+01 $\pm$ 2.31E+01 \\\hline
MABC  & -5.61E+02 $\pm$ 1.38E+00 & -2.82E+02 $\pm$ 3.27E+01 &\bf{-1.94E+02 $\pm$ 1.33E+01} & -7.01E+01 $\pm$ 1.05E+01 \\\hline
ABC  & -5.61E+02 $\pm$ 1.44E+00 & -2.37E+02 $\pm$ 3.51E+01 & -1.84E+02 $\pm$ 1.78E+01 & -6.84E+01 $\pm$ 1.41E+01 \\\hline
DE & -5.61E+02 $\pm$ 1.10E+00 & -4.30E+01 $\pm$ 1.05E+02 & -1.59E+02 $\pm$ 2.11E+01 & -9.86E+00 $\pm$ 1.22E+01  \\\hline
WOA  & -5.64E+02 $\pm$ 2.78E+00 & -4.45E+02 $\pm$ 2.21E+01 & 7.52E+01 $\pm$ 9.86E+01 & 1.46E+02 $\pm$ 1.03E+02 \\\hline
SCA  & -5.61E+02 $\pm$ 1.21E+00 & 1.01E+03 $\pm$ 3.28E+02 & -4.43E+01 $\pm$ 2.90E+01 & 7.99E+01 $\pm$ 3.52E+01 \\\hline
  Algorithm &  F13 &  F14  &  F15  &  F16   \\\hline
ASBSO   & 1.30E+02 $\pm$ 6.54E+01 &\bf{ 3.68E+03 $\pm$ 4.56E+02} & 3.88E+03 $\pm$ 5.74E+02 & 2.00E+02 $\pm$ 1.13E-01 \\\hline
CGSA-M  & 2.61E+02 $\pm$ 3.66E+01 & 3.89E+03 $\pm$ 4.67E+02 & \bf{3.78E+03 $\pm$ 4.89E+02} & 2.00E+02 $\pm$ 4.63E-03 \\\hline
MABC  & \bf{1.80E+01 $\pm$ 1.40E+01} & 7.11E+03 $\pm$ 2.23E+02 & 7.45E+03 $\pm$ 2.00E+02 & 2.02E+02 $\pm$ 2.54E-01 \\\hline
ABC  & 2.24E+01 $\pm$ 9.10E+00 & 7.15E+03 $\pm$ 2.18E+02 & 7.46E+03 $\pm$ 2.37E+02 & 2.02E+02 $\pm$ 2.93E-01 \\\hline
DE & 9.85E+01 $\pm$ 8.77E+00 & 6.61E+03 $\pm$ 4.55E+02 & 7.47E+03 $\pm$ 2.43E+02 & 2.02E+02 $\pm$ 3.33E-01  \\\hline
WOA  & 2.99E+02 $\pm$ 8.96E+01 & 4.88E+03 $\pm$ 7.84E+02 & 5.49E+03 $\pm$ 1.02E+03 & 2.02E+02 $\pm$ 4.34E-01  \\\hline
SCA  & 1.67E+02 $\pm$ 3.69E+01 & 7.00E+03 $\pm$ 3.40E+02 & 7.49E+03 $\pm$ 2.06E+02 & 2.02E+02 $\pm$ 2.55E-01  \\ \hline
  Algorithm &  F17  &  F18  &  F19  &  F20   \\\hline
ASBSO & 5.28E+02 $\pm$ 5.13E+01 & 5.98E+02 $\pm$ 2.85E+01 & \bf{5.04E+02 $\pm$ 7.60E-01} & 6.14E+02 $\pm$ 2.94E-01  \\\hline
CGSA-M  &\bf{ 3.66E+02 $\pm$ 8.04E+00} & \bf{4.55E+02 $\pm$ 5.69E+00} & 5.11E+02 $\pm$ 2.40E+00 & 6.15E+02 $\pm$ 2.23E-01  \\\hline
MABC  & 5.31E+02 $\pm$ 1.23E+01 & 6.42E+02 $\pm$ 1.05E+01 & 1.50E+03 $\pm$ 5.36E+02 & 6.15E+02 $\pm$ 1.25E-01 \\\hline
ABC  & 5.38E+02 $\pm$ 1.08E+01 & 6.44E+02 $\pm$ 8.95E+00 & 1.74E+03 $\pm$ 5.62E+02 & 6.15E+02 $\pm$ 1.37E-01  \\\hline
DE & 6.31E+02 $\pm$ 3.76E+01 & 7.47E+02 $\pm$ 3.22E+01 & 5.37E+02 $\pm$ 1.24E+01 &\bf{ 6.13E+02 $\pm$ 1.37E-01} \\\hline
WOA  & 8.89E+02 $\pm$ 1.11E+02 & 1.01E+03 $\pm$ 1.20E+02 & 5.58E+02 $\pm$ 1.90E+01 & 6.15E+02 $\pm$ 3.05E-01  \\\hline
SCA  & 7.88E+02 $\pm$ 4.68E+01 & 8.88E+02 $\pm$ 3.94E+01 & 2.99E+03 $\pm$ 1.30E+03 & 6.14E+02 $\pm$ 3.54E-01  \\\hline
  Algorithm &  F21  &  F22  &  F23  &  F24   \\\hline
ASBSO & 1.02E+03 $\pm$ 8.42E+01 & \bf{5.36E+03 $\pm$ 4.70E+02} & \bf{6.00E+03 $\pm$ 7.84E+02} & 1.31E+03 $\pm$ 2.60E+01 \\\hline
CGSA-M  & 1.01E+03 $\pm$ 4.38E+01 & 7.42E+03 $\pm$ 5.68E+02 & 6.81E+03 $\pm$ 3.14E+02 & 1.35E+03 $\pm$ 6.98E+01  \\\hline
MABC  & \bf{9.94E+02 $\pm$ 2.19E+01} & 8.69E+03 $\pm$ 2.71E+02 & 8.84E+03 $\pm$ 3.22E+02 & \bf{1.28E+03 $\pm$ 7.18E+00}  \\\hline
ABC  & 1.00E+03 $\pm$ 1.33E-02 & 8.74E+03 $\pm$ 2.01E+02 & 8.82E+03 $\pm$ 3.04E+02 & 1.29E+03 $\pm$ 5.63E+00 \\\hline
DE & 1.57E+03 $\pm$ 1.87E+02 & 7.76E+03 $\pm$ 4.13E+02 & 8.44E+03 $\pm$ 3.02E+02 & 1.30E+03 $\pm$ 2.66E+00 \\\hline
WOA  & 1.03E+03 $\pm$ 6.87E+01 & 6.76E+03 $\pm$ 1.08E+03 & 7.61E+03 $\pm$ 8.55E+02 & 1.31E+03 $\pm$ 1.00E+01  \\\hline
SCA  & 2.58E+03 $\pm$ 1.79E+02 & 8.35E+03 $\pm$ 4.43E+02 & 8.70E+03 $\pm$ 3.72E+02 & 1.32E+03 $\pm$ 5.05E+00 \\ \hline
  Algorithm &  F25  &  F26  &  F27  &  F28   \\\hline
ASBSO& \bf{1.41E+03 $\pm$ 1.03E+01} & 1.46E+03 $\pm$ 7.82E+01 & 2.42E+03 $\pm$ 1.07E+02 & 2.03E+03 $\pm$ 8.02E+02 \\\hline
CGSA-M& 1.49E+03 $\pm$ 7.05E+00 & 1.55E+03 $\pm$ 3.14E+01 & \bf{2.23E+03 $\pm$ 8.05E+01} & 5.00E+03 $\pm$ 2.48E+02 \\\hline
MABC & 1.44E+03 $\pm$ 3.82E+00 & 1.42E+03 $\pm$ 5.30E+00 & 2.66E+03 $\pm$ 4.42E+01 & \bf{1.70E+03 $\pm$ 9.31E-05} \\\hline
ABC & 1.44E+03 $\pm$ 5.31E+00 & 1.42E+03 $\pm$ 6.02E+00 & 2.67E+03 $\pm$ 4.60E+01 & 1.70E+03 $\pm$ 1.47E+00 \\\hline
DE & 1.42E+03 $\pm$ 3.38E+00 & \bf{1.41E+03 $\pm$ 2.06E+00} & 2.63E+03 $\pm$ 2.62E+01 & 2.66E+03 $\pm$ 1.29E+02 \\\hline
WOA& 1.42E+03 $\pm$ 9.66E+00 & 1.53E+03 $\pm$ 9.64E+01 & 2.61E+03 $\pm$ 6.95E+01 & 5.36E+03 $\pm$ 7.53E+02 \\\hline
SCA & 1.43E+03 $\pm$ 4.21E+00 & 1.41E+03 $\pm$ 5.66E+00 & 2.66E+03 $\pm$ 4.51E+01 & 3.92E+03 $\pm$ 1.98E+02 \\\hline
\end{tabular}}
\label{dataF1F28}
\end{table*}

\begin{table*}[!htbp]
\caption{Experimental results of CEC'17 (F29-F57) using ASBSO, CGSA-M, MABC, ABC, DE, WOA and SCA.}
\centering
\scalebox{1.0}{
\begin{tabular}{lcccc}\\\hline
Algorithm &  F29  &  F30  &  F31  &  F32   \\\hline
ASBSO & 2.21E+03 $\pm$ 2.00E+03 & \bf{3.95E+02 $\pm$ 1.10E+02} & \bf{4.72E+02 $\pm$ 2.92E+01} & \bf{6.86E+02 $\pm$ 3.45E+01}  \\\hline
CGSA-M  &\bf{ 1.82E+03 $\pm$ 9.25E+02} & 8.51E+04 $\pm$ 6.02E+03 & 5.34E+02 $\pm$ 1.24E+01 & 7.33E+02 $\pm$ 1.99E+01 \\\hline
MABC  & 2.01E+03 $\pm$ 1.82E+03 & 9.71E+04 $\pm$ 1.28E+04 & 5.17E+02 $\pm$ 2.31E+00 & 7.19E+02 $\pm$ 1.48E+01  \\\hline
ABC  & 4.72E+04 $\pm$ 7.74E+04 & 1.03E+05 $\pm$ 1.09E+04 & 5.19E+02 $\pm$ 2.79E+00 & 7.18E+02 $\pm$ 9.44E+00  \\\hline
DE & 1.32E+09 $\pm$ 4.55E+08 & 8.04E+04 $\pm$ 1.05E+04 & 6.25E+02 $\pm$ 2.90E+01 & 7.50E+02 $\pm$ 1.31E+01  \\\hline
WOA  & 2.48E+06 $\pm$ 1.73E+06 & 1.60E+05 $\pm$ 8.12E+04 & 5.45E+02 $\pm$ 3.70E+01 & 7.64E+02 $\pm$ 6.16E+01  \\\hline
SCA  & 1.18E+10 $\pm$ 1.77E+09 & 3.50E+04 $\pm$ 6.31E+03 & 1.40E+03 $\pm$ 2.74E+02 & 7.71E+02 $\pm$ 2.17E+01  \\\hline
  Algorithm &  F33  &  F34  &  F35  &  F36   \\\hline
ASBSO & 6.51E+02 $\pm$ 7.77E+00 & 1.16E+03 $\pm$ 9.94E+01 &\bf{ 9.41E+02 $\pm$ 3.19E+01} & 3.93E+03 $\pm$ 6.39E+02  \\\hline
CGSA-M  & 6.50E+02 $\pm$ 3.78E+00 & \bf{7.86E+02 $\pm$ 1.11E+01} & 9.52E+02 $\pm$ 1.34E+01 & 2.96E+03 $\pm$ 2.48E+02 \\\hline
MABC  & \bf{6.00E+02 $\pm$ 7.55E-04} & 9.39E+02 $\pm$ 9.74E+00 & 1.02E+03 $\pm$ 1.08E+01 & \bf{1.41E+03 $\pm$ 3.36E+02}  \\\hline
ABC  & 6.00E+02 $\pm$ 6.69E-03 & 9.43E+02 $\pm$ 9.71E+00 & 1.02E+03 $\pm$ 1.16E+01 & 1.90E+03 $\pm$ 4.45E+02  \\\hline
DE & 6.24E+02 $\pm$ 4.43E+00 & 1.17E+03 $\pm$ 9.96E+01 & 1.06E+03 $\pm$ 1.14E+01 & 4.14E+03 $\pm$ 7.85E+02  \\\hline
WOA  & 6.67E+02 $\pm$ 1.12E+01 & 1.21E+03 $\pm$ 9.33E+01 & 9.99E+02 $\pm$ 3.20E+01 & 6.54E+03 $\pm$ 2.35E+03 \\\hline
SCA  & 6.49E+02 $\pm$ 5.34E+00 & 1.12E+03 $\pm$ 2.88E+01 & 1.05E+03 $\pm$ 1.63E+01 & 5.52E+03 $\pm$ 1.10E+03 \\\hline
  Algorithm &  F37  &  F38  &  F39  &  F40   \\\hline
ASBSO & 5.20E+03 $\pm$ 5.67E+02 & \bf{1.23E+03 $\pm$ 4.75E+01} & \bf{1.41E+06 $\pm$ 8.00E+05} & 5.04E+04 $\pm$ 2.64E+04 \\\hline
 CGSA-M & \bf{4.83E+03 $\pm$ 4.17E+02} & 1.46E+03 $\pm$ 7.26E+01 & 1.49E+07 $\pm$ 2.37E+07 & 3.02E+04 $\pm$ 5.39E+03  \\\hline
 MABC & 8.15E+03 $\pm$ 3.09E+02 & 4.31E+03 $\pm$ 6.19E+02 & 7.79E+07 $\pm$ 2.79E+07 & 8.46E+07 $\pm$ 2.90E+07 \\\hline
ABC  & 8.10E+03 $\pm$ 3.19E+02 & 4.37E+03 $\pm$ 7.31E+02 & 1.17E+08 $\pm$ 2.66E+07 & 8.02E+07 $\pm$ 3.32E+07 \\\hline
DE & 8.17E+03 $\pm$ 2.51E+02 & 1.33E+03 $\pm$ 2.16E+01 & 5.43E+07 $\pm$ 1.60E+07 & \bf{4.13E+03 $\pm$ 5.37E+02}  \\\hline
WOA  & 6.06E+03 $\pm$ 9.74E+02 & 1.45E+03 $\pm$ 1.15E+02 & 4.47E+07 $\pm$ 3.11E+07 & 1.35E+05 $\pm$ 1.44E+05  \\\hline
SCA  & 8.12E+03 $\pm$ 3.34E+02 & 2.19E+03 $\pm$ 3.99E+02 & 1.21E+09 $\pm$ 2.30E+08 & 4.07E+08 $\pm$ 1.98E+08 \\\hline
  Algorithm &  F41 &  F42  &  F43  &  F44   \\\hline
ASBSO & 7.08E+03 $\pm$ 5.23E+03 & 3.01E+04 $\pm$ 2.25E+04 & \bf{3.01E+03 $\pm$ 2.25E+02} & \bf{2.40E+03 $\pm$ 2.44E+02}  \\\hline
CGSA-M  & 4.79E+05 $\pm$ 1.35E+05 & 1.21E+04 $\pm$ 1.59E+03 & 3.16E+03 $\pm$ 2.60E+02 & 2.81E+03 $\pm$ 2.33E+02  \\\hline
MABC  & 3.62E+05 $\pm$ 1.64E+05 & 1.96E+07 $\pm$ 7.41E+06 & 3.68E+03 $\pm$ 1.57E+02 & 2.50E+03 $\pm$ 1.17E+02 \\\hline
ABC  & 3.04E+05 $\pm$ 1.25E+05 & 2.08E+07 $\pm$ 8.65E+06 & 3.76E+03 $\pm$ 1.87E+02 & 2.49E+03 $\pm$ 1.19E+02  \\\hline
DE & \bf{1.49E+03 $\pm$ 7.57E+00} & \bf{1.72E+03 $\pm$ 3.06E+01} & 3.19E+03 $\pm$ 3.08E+02 & 2.41E+03 $\pm$ 2.16E+02  \\\hline
WOA  & 7.27E+05 $\pm$ 6.79E+05 & 6.97E+04 $\pm$ 4.48E+04 & 3.47E+03 $\pm$ 5.17E+02 & 2.53E+03 $\pm$ 2.16E+02 \\\hline
SCA  & 1.19E+05 $\pm$ 7.05E+04 & 1.56E+07 $\pm$ 1.29E+07 & 3.64E+03 $\pm$ 2.13E+02 & 2.42E+03 $\pm$ 1.64E+02 \\\hline
  Algorithm &  F45  &  F46  &  F47  &  F48   \\\hline
ASBSO & 1.23E+05 $\pm$ 1.21E+05 & 1.25E+05 $\pm$ 6.31E+04 & 2.67E+03 $\pm$ 2.17E+02 & \bf{2.49E+03 $\pm$ 3.14E+01}  \\\hline
CGSA-M  & 2.99E+05 $\pm$ 1.33E+05 & 1.56E+04 $\pm$ 5.55E+03 & 3.01E+03 $\pm$ 2.02E+02 & 2.56E+03 $\pm$ 2.59E+01  \\\hline
 MABC & 6.77E+06 $\pm$ 2.63E+06 & 2.67E+07 $\pm$ 1.04E+07 & 2.75E+03 $\pm$ 1.06E+02 & 2.51E+03 $\pm$ 1.18E+01  \\\hline
ABC  & 6.34E+06 $\pm$ 3.20E+06 & 2.39E+07 $\pm$ 1.03E+07 & 2.74E+03 $\pm$ 8.15E+01 & 2.52E+03 $\pm$ 1.18E+01 \\\hline
DE & \bf{6.90E+03 $\pm$ 1.84E+03} &\bf{ 1.96E+03 $\pm$ 4.88E+00} & \bf{2.31E+03 $\pm$ 2.03E+02} & 2.54E+03 $\pm$ 1.26E+01  \\\hline
WOA  & 3.02E+06 $\pm$ 2.58E+06 & 2.45E+06 $\pm$ 2.03E+06 & 2.78E+03 $\pm$ 1.76E+02 & 2.56E+03 $\pm$ 6.24E+01 \\\hline
SCA  & 2.80E+06 $\pm$ 1.21E+06 & 2.48E+07 $\pm$ 1.13E+07 & 2.61E+03 $\pm$ 1.29E+02 & 2.56E+03 $\pm$ 1.92E+01  \\\hline
  Algorithm &  F49  &  F50  &  F51  &  F52   \\\hline
ASBSO & 5.79E+03 $\pm$ 2.04E+03 & 3.26E+03 $\pm$ 1.24E+02 & 3.49E+03 $\pm$ 9.56E+01 & 2.89E+03 $\pm$ 1.25E+01  \\\hline
CGSA-M  & 6.20E+03 $\pm$ 1.84E+03 & 3.61E+03 $\pm$ 1.59E+02 & 3.27E+03 $\pm$ 5.85E+01 & 2.93E+03 $\pm$ 1.24E+01 \\\hline
MABC  & 2.52E+03 $\pm$ 1.76E+02 & 2.88E+03 $\pm$ 1.65E+01 & 3.04E+03 $\pm$ 1.19E+01 & \bf{2.89E+03 $\pm$ 1.29E-01}  \\\hline
ABC  & 2.64E+03 $\pm$ 2.08E+02 & 2.89E+03 $\pm$ 1.60E+01 &3.04E+03 $\pm$ 1.17E+01 & 2.89E+03 $\pm$ 1.73E-01  \\\hline
DE & \bf{2.52E+03 $\pm$ 4.78E+01} &\bf{ 2.88E+03 $\pm$ 1.38E+01} &  \bf{3.04E+03 $\pm$ 1.07E+01} & 3.01E+03 $\pm$ 3.38E+01 \\\hline
WOA  & 6.65E+03 $\pm$ 1.87E+03 & 3.05E+03 $\pm$ 8.54E+01 & 3.16E+03 $\pm$ 9.15E+01 & 2.94E+03 $\pm$ 2.73E+01 \\\hline
SCA  & 8.25E+03 $\pm$ 2.37E+03 & 2.99E+03 $\pm$ 2.34E+01 & 3.16E+03 $\pm$ 2.96E+01 & 3.20E+03 $\pm$ 4.91E+01  \\\hline
 Algorithm &   F53  &  F54  &  F55  &  F56     \\\hline
ASBSO  & 7.84E+03 $\pm$ 1.80E+03 & 3.85E+03 $\pm$ 2.17E+02 & \bf{3.18E+03 $\pm$ 3.60E+01} &4.40E+03 $\pm$ 3.39E+02  \\\hline
CGSA-M  & 6.75E+03 $\pm$ 6.48E+02 & 4.51E+03 $\pm$ 3.18E+02 & 3.31E+03 $\pm$ 5.88E+01 & 4.71E+03 $\pm$ 2.26E+02  \\\hline
MABC  & 5.71E+03 $\pm$ 1.13E+02 & 3.46E+03 $\pm$ 2.89E+01 & 3.23E+03 $\pm$ 1.95E+01 & 4.86E+03 $\pm$ 1.75E+02  \\\hline
ABC & 5.74E+03 $\pm$ 1.28E+02 & 3.46E+03 $\pm$ 3.87E+01 & 3.26E+03 $\pm$ 2.59E+01 & 4.93E+03 $\pm$ 1.31E+02  \\\hline
DE & \bf{3.04E+03 $\pm$ 1.07E+01} & \bf{3.01E+03 $\pm$ 3.38E+01} & 6.15E+03 $\pm$ 1.47E+02 & \bf{3.26E+03 $\pm$ 1.20E+01} \\\hline
WOA & 7.24E+03 $\pm$ 1.00E+03 & 3.36E+03 $\pm$ 8.88E+01 & 3.31E+03 $\pm$ 3.96E+01 & 5.00E+03 $\pm$ 4.58E+02  \\\hline
SCA& 6.87E+03 $\pm$ 2.56E+02 & 3.39E+03 $\pm$ 4.83E+01 & 3.78E+03 $\pm$ 1.36E+02 & 4.62E+03 $\pm$ 2.62E+02  \\ \hline
Algorithm &  F57    \\\cline{1-2}
ASBSO& 5.16E+05 $\pm$ 2.90E+05\\ \cline{1-2}
CGSA-M& \bf{1.48E+05 $\pm$ 8.05E+04}\\ \cline{1-2}
MABC& 2.38E+07 $\pm$ 7.73E+06\\ \cline{1-2}
ABC& 2.67E+07 $\pm$ 1.04E+07\\\cline{1-2}
DE & 1.94E+05 $\pm$ 7.07E+04 \\\cline{1-2}
WOA& 1.04E+07 $\pm$ 6.42E+06\\ \cline{1-2}
SCA& 7.44E+07 $\pm$ 2.59E+07\\ \cline{1-2}
\end{tabular}}
\label{dataF29F56}
\end{table*}

\section{Experimental Results}
\label{section4}
Two groups of comparisons have been carried out which include internal comparisons and external comparisons using CEC'13 and CEC'17 test functions. It should be noticed that F2 in CEC'17 has been excluded because it shows unstable behavior especially for higher dimensions, and significant performance variations for the same algorithm implemented in Matlab, or C Language \cite{liang2013problem,awad2016problem}. The internal comparison aims to demonstrate that ASBSO can achieve better performance than BSO not only at low dimension, but also at high dimension. Therefore, these comprehensive comparisons can show the search ability and robustness of ASBSO for solving the problems with different difficulty levels.

After proving the superiority of ASBSO, in the external comparison, some meta-heuristic algorithms have been taken into account to further evaluate the performance of ASBSO. Artificial bee colony algorithm (ABC) \cite{tereshko2005collective} is very popular in literature and its influence is next only to particle swarm optimization (PSO) \cite{kennedy2011particle} in swarm-based meta-heuristic algorithms \cite{zhang2015comprehensive,shi2016improved,zhang2017multi}. Differential evolution (DE) \cite{storn1997differential,gao2017understanding,parouha2016robust} is the most famous optimization algorithm with very powerful search ability. MABC and CGSA-M \cite{li2016artificial,song2017multiple} which are two variations based on ABC and gravitational search algorithm (GSA) \cite{rashedi2009gsa,ji2017self,sun2018stability} implement memory-based selection strategies. Thus, they are very suitable to be chosen to compare with ASBSO. Furthermore, two newly proposed effective swarm intelligence based algorithms, i.e., whale optimization algorithm (WOA) \cite{mirjalili2016whale} and sine cosine algorithm (SCA) \cite{mirjalili2016sca}, have been implemented.  The population size of all compared algorithms is 100. All these contrast experiments are run for 30 times to reduce the random error, and the maximum number of function evaluation is set to 10000$D$ ($D$ is the dimension number).

\subsection{Parameter Analysis}
The aim of implementing multiple strategies and memory based selection method is to provide multiple step lengths in order to suit different search phases. Too few strategies couldn't satisfy this demand while too many strategies are redundant and will increase computational cost. Thus, we attempt $M=4$ in this paper and preliminary experiments prove the validity of this parameter setting. A parameter analysis is executed to find an applicable value for $H$. Three values are applied involving 10, 20 and 30. In this comparison, $k$ is set to 10. The contrast experiment is implement on CEC'13 and CEC'17 to find the most suitable value for four strategies.

Friedman test for multiple comparison is applied to analyze the results \cite{luengo2009study}.  Table \ref{h} lists statistical results obtained by Friedman test and $H=20$ is the control algorithm. $Ranking$ evaluates the performance of each algorithm, and a lower ranking indicates a better performance. Unadjusted $p$-value doesn't consider the probability error in a multiple comparison. Thus, two commonly used post-hoc procedures, Holm and Hochberg procedures \cite{garcia2009study}, are taken into account  and their conservative adjusted $p$-values are convincing enough to eliminate Type I error \cite{garcia2010advanced}. $H=20$ which maintains the best ranking of 1.4298 indicates that it's the best value for $H$. Therefore, $k=10$ and $H=20$ are chosen to be applied into the flexible multiple search length strategy.

\subsection{Internal Comparison}
In the first experiment, the CEC'13 and CEC'17 are used to compare the performance between traditional BSO and the proposed ASBSO. The experiments are tested at dimension $D=10, 30, 50$ and 100 respectively.

The experimental results of CEC'13 are summarized in Tables \ref{D10D30} and \ref{D50D100},  while Tables \ref{17D10D30} and \ref{17D50D100} show the results of CEC'17. All the better Mean and standard deviation (Std Dev) values are highlighted for convenience. From these tables, we can intuitively find out that ASBSO can obtain more number of better results than BSO. The former obtains better results on F4, F7, F9, F11, F14, F15, F17-F19, F27 and F28 at all tested dimensions, while BSO only obtains better result on F6 in CEC'13. In CEC'17, ASBSO outperforms on F29, F32, F43 and F46 while BSO can't obtain better performance at all dimensions on any function.

Wilcoxon signed-rank test is conducted to prove that ASBSO can beat BSO as it's a pairwise test which is used to analyze significant difference between the performance of two algorithms. $R^{+}$ and $R^{-}$ values in Tables \ref{wilcoxon13} and \ref{wilcoxon17} can indicate the degree that ASBSO outperforms BSO. As we conduct ASBSO versus BSO, $R^{+}$ represents the sum of ranks for the functions on which ASBSO outperforms BSO, and $R^{-}$ means the opposite.With the null hypothesis $H_0$ for the test assumes two compared algorithms have no difference,  a better performance of our proposed algorithm can be shown via a higher $R^{+}$ value and $p$-value indicates the possibility that the null hypothesis happens. If $p$-value is lower than the level of significance $\alpha=0.05$, we can accept the hypothesis that ASBSO is significantly better than BSO. Moreover, we set a more rigorous level $\alpha=0.01$ to further exhibit the improvement of ASBSO in solution quality.

All the comparisons in Table \ref{wilcoxon13} can reach the level of $\alpha=0.01$, while in Table \ref{wilcoxon17}, ASBSO can beat BSO on the level of  $\alpha=0.05$ at all dimensions but only has a significant difference at $D=100$ when  $\alpha=0.01$. It's understandable because CEC'17 is a newly proposed benchmark function suit, all test functions have a promotion in difficulty and complexity compared with CEC'13.

From these results, it can be concluded that ASBSO has obvious advantage in comparison with BSO in terms of search ability and solution quality.

\subsection{External Comparison}
To investigate the performance of ASBSO when comparing with other swarm intelligence optimization algorithms,  some well-known meta-heuristic algorithms, involving CGSA-M, MABC, ABC, DE, WOA and SCA, are implemented into numerical tests. Parameter settings can be investigated according to \cite{song2017multiple, li2016artificial, tereshko2005collective, mirjalili2016whale, mirjalili2016sca}. In DE, we use the efficient parameter set $F=0.9$ and $CR=0.9$ as suggested in \cite{brest2006self, liu2005fuzzy}. All tests have been executed at $D=30$ with maximum number of function evaluation equals 10000$D$ for 30 runs.

The results are listed in Tables \ref{dataF1F28} and \ref{dataF29F56}. The best results are marked in boldface. It's visual that ASBSO obtains the largest number of the best results among all compared algorithms and we can draw a preliminary conclusion that ASBSO is very competitive in contrast with others. To more precisely analyze the results of multiple comparisons, Friedman test \cite{luengo2009study} which is widely used in \cite{cai2013differential,wang2014differential,wang2016cooperative} is employed. Table \ref{friedman57} lists statistical results obtained by Friedman test and ASBSO is the control algorithm. ASBSO maintains the best ranking of 2.5 while the second best is only 3.5526 which belongs to CGSA-M. Although adjusted $p$-values of Holm and Hochberg procedures are multiplied bigger than unadjusted $p$-values, they still reach the significant level of $\alpha=0.05$. Furthermore, in terms of MABC, ABC, WOA and SCA, adjusted $p$-values satisfy the level of $\alpha=0.01$. Wilcoxon test is also conducted to verify the results of Friedman test and obtains similar $p$-values in Table \ref{wilcoxon57}. From all these results, it is obvious that ASBSO is significantly better than other contrast algorithms in benchmark function tests.

\begin{table*}[!htbp]
\caption{Adjusted $p$-values (FRIEDMAN).}
\centering
\scalebox{1.0}{
\begin{tabular}{ccccccc}\\\hline
Algorithm & Ranking&unadjusted $p$&$p_{Holm}$&$p_{Hochberg}$ & $\alpha=0.05$& $\alpha=0.01$ \\\hline
ASBSO vs.&2.5  \\\hline
CGSA-M&3.5526&0.009286&0.011049&0.009286&YES&NO\\\hline
MABC& 3.9737&0.000271&0.000812&0.000812&YES&YES\\\hline
ABC&4.3509&0.000005&0.000019&0.000019&YES&YES\\\hline
DE&3.6228&0.005524&0.011049&0.009286&YES&NO\\\hline
WOA&4.7982 &0&0&0&YES&YES\\\hline
SCA&5.2018 &0&0&0&YES&YES\\\hline
\label{friedman57}
\end{tabular}}
\end{table*}

\begin{table*}[!htbp]
\caption{Results obtained by the Wilcoxon signed-rank test for ASBSO vs. some other typical algorithms.}
\centering
\scalebox{1.0}{
\begin{tabular}{ccccccc}\\\hline
Algorithm & $R^{+}$ & $R^{-}$ & $p$-value & $\alpha=0.05$& $\alpha=0.01$ \\\hline
ASBSO vs.  \\\hline
CGSA-M& 1046.5 & 549.5  & 4.1615E-2&YES&NO\\\hline
MABC& 1208.5 & 387.5 &  7.81E-4 &YES&YES\\\hline
ABC   & 1256.0 & 340.0 & 1.73E-4&YES&YES\\\hline
DE & 1040.5 & 555.5 &4.195E-2&YES&NO\\ \hline
WOA & 1473.0 & 123.0 & 0.00 &YES&YES\\\hline
SCA  &1510.0 & 143.0 & 0.00&YES&YES\\\hline
\label{wilcoxon57}
\end{tabular}}
\end{table*}

\begin{table*}[!htbp]
\caption{Experimental results on real-world problems.}
\centering
\scalebox{0.95}{
\begin{tabular}{l|c|c|c|c}\hline
                           & BSO                                           &  ASBSO                                     & CGSA-M                               &MABC                                                      \\\hline
RF1                      &1.30E+01$\pm$4.98E+00    &\bf{9.93E+00$\pm$4.68E+00} &  2.19E+01$\pm$4.33E+00  &1.81E+01$\pm$2.13E+00\\\hline
RF2                      & -2.11E+01$\pm$	2.95E+00& \bf{-2.52E+01$\pm$2.12E+00} &-3.24E+00$\pm$1.08E+00& -1.17E+01$\pm$9.23E-01  \\\hline
RF4                      & 1.50E+01$\pm$9.12E-01 & \bf{1.47E+01$\pm$5.15E-01}  &1.99E+01$\pm$2.09E+00 & 1.65E+01$\pm$2.37E+00    \\\hline
RF7                      & 8.72E-01$\pm$1.01E-01&\bf{7.91E-01$\pm$1.49E-01}      &2.55E+00$\pm$2.39E-01      &1.63E+00$\pm$9.01E-02     \\\hline
                             &ABC                     & DE                                      & WOA                                 &SCA\\\hline
RF1                     &1.83E+01$\pm$1.65E+00    &3.15E+01$\pm$2.01E+01    &2.19E+01$\pm$5.00E+00 &  1.83E+01$\pm$3.98E+00\\\hline
RF2                      &-1.13E+01$\pm$5.04E-01&-4.24E+00$\pm$4.02E-01& -1.84E+01$\pm$4.89E+00 &-9.20E+00$\pm$1.16E+00 \\\hline
RF4                      &1.63E+01$\pm$2.04E+00& 4.99E+01$\pm$2.46E+01 &1.47E+01$\pm$1.26E+00 &1.51E+01$\pm$1.19E+00  \\\hline
RF7                      &1.67E+00$\pm$8.83E-02& 3.31E+00$\pm$4.06E-01&1.92E+00$\pm$2.30E-01  &2.12E+00$\pm$2.02E-01  \\\hline
\end{tabular}}
\label{realworld}
\end{table*}

To visually demonstrate the comparisons among ASBSO and other contrast algorithms, six functions, F4, F14, F22, F39, F43 and F48 with different properties, including unimodal, simple multimodal, hybrid and composition, are selected since they are representative to show the properties of all tested functions. The convergent procedures and final solutions obtained by these algorithms in all 30 runs are exhibited.

Fig. \ref{fig2} is the box-and-whisker diagrams and Fig. \ref{fig3} is the convergence graphs. Five values including median, maximum, minimum, first quartile and third quartile are shown in box-and-whisker plots. The range between the first quartile and the third quartile is called interquartile range (IQR), and if the points locate either 1.5*IQR above the third quartile (i.e. 1.5*IQR) below the first quartile, they are marked as outliers. Extreme outliers refer to the points locate either 3*IQR above the third quartile or 3*IQR below the first quartile. In these six plots, the median values of ASBSO are the smallest and its IQRs are lower and shorter than most other algorithms. These indicate that the solution quality and stability obtained by ASBSO is much better than those of other contrast algorithms.

The convergence graphs can not only demonstrate the precision of solutions but also compare the convergence speeds. Fig. \ref{fig3} shows that, ASBSO can possess the fast convergence speed. In details, all algorithms' convergence behaviors shown in Fig. \ref{fig3} (a) are quite illuminating to further elaborate the search behavior of ASBSO. It is clear that ASBSO continues converging when other algorithms stop in the latter of the search iteration. Although ABC starts with a better initial position, it doesn't have the ability to jump out of local optima and ultimately be transcended by ASBSO. In the comparison between ASBSO and BSO, it illustrates that the former always has a better solution precision and convergence speed than the latter. When comparing with other algorithms, ASBSO also obtains fabulous performances. Thus, it can be concluded that the proposed adaptive step length based on memory selection method enhances the search ability and efficiency for ASBSO.

\subsection{Real World Optimization Problems}
It has been demonstrated that ASBSO can outperform traditional BSO and other well-known algorithms on benchmark functions. To further testify its application value, four problems introduced in CEC'11 \cite{das2010problem} are used to execute this test: (1) RF1: Parameter Estimation for Frequency-Modulated (FM) Sound Waves, (2) RF2: Lennard-Jones Potential Problem, (3) RF4: Optimal Control of a Non-Linear Stirred Tank Reactor, and (4) RF7: Transmission Network Expansion Planning (TNEP) problem \cite{das2010problem}. All these problems are run for 30 independent times and the maximum function evaluation is set to $10000D$. The experimental results are presented in Table \ref{realworld}. It's obvious that  ASBSO obtains dominance over all tested problems when compared with other algorithms, which well exhibiting its application value.

\begin{table*}[!htbp]
\centering
\caption{Experimental results  of using ASBSO and  BSO with $1/5$ Success Rule on CEC'13 and CEC'17 benchmark functions (F1-F57).}
\begin{tabular}{c|c|c|c|c|c}\hline
    \multicolumn{6}{c}{$D$=30}                                                                            \\\hline
  &ASBSO                                           &BSO with $1/5$ Rule                               &   & ASBSO                       &BSO with $1/5$ Rule \\\hline
  & Mean (Std Dev)                        & Mean (Std Dev)                   &     & Mean (Std Dev)             & Mean (Std Dev)  \\\hline
F1	&	-1.40E+03	(	1.98E-13	)	&	-1.40E+03	(	7.26E-13	)	&	F29	&	\bf{2.21E+03	(	2.00E+03	)}	&	2.93E+03	(	3.12E+03	)\\\hline
F2	&	\bf{1.54E+06	(	4.26E+05	)}	&	3.96E+06	(	1.25E+06	)	&	F30	&	\bf{3.95E+02	(	1.10E+02	)}	&	5.39E+02	(	2.13E+02	)\\\hline
F3	&	\bf{8.47E+07	(	8.64E+07	)}	&	4.28E+08	(	4.44E+08	)	&	F31	&	\bf{4.72E+02	(	2.92E+01	)}	&	5.00E+02	(	1.92E+01	)\\\hline
F4	&	5.08E+03	(	2.25E+03	)	&	\bf{1.86E+03	(	1.34E+03	)}	&	F32	&	6.86E+02	(	3.45E+01	)	&	\bf{6.79E+02	(	2.61E+01	)}\\\hline
F5	&	-1.00E+03	(	2.98E-03	)	&	-1.00E+03	(	3.99E-04	)	&	F33	&	\bf{6.51E+02	(	7.77E+00	)}	&	6.52E+02	(	9.75E+00	)\\\hline
F6	&	\bf{-8.64E+02	(	2.46E+01	)}	&	-8.50E+02	(	2.90E+01	)	&	F34	&	1.16E+03	(	9.94E+01	)	&	\bf{1.11E+03	(	1.01E+02	)}\\\hline
F7	&	\bf{-7.08E+02	(	3.90E+01	)}	&	-6.52E+02	(	4.02E+01	)	&	F35	&	\bf{9.41E+02	(	3.19E+01	)}	&	9.35E+02	(	2.82E+01	)\\\hline
F8	&	-6.79E+02	(	6.70E-02	)	&	-6.79E+02	(	9.75E-02	)	&	F36	&	\bf{3.93E+03	(	6.39E+02	)}	&	4.59E+03	(	8.43E+02	)\\\hline
F9	&	\bf{-5.71E+02	(	3.23E+00	)}	&	-5.65E+02	(	2.53E+00	)	&	F37	&	\bf{5.20E+03	(	5.67E+02	)}	&	5.78E+03	(	9.50E+02	)\\\hline
F10	&	\bf{-5.00E+02	(	5.35E-02	)}	&	-4.99E+02	(	4.49E-01	)	&	F38	&	\bf{1.23E+03	(	4.75E+01	)}	&	1.26E+03	(	5.13E+01	)\\\hline
F11	&	\bf{-1.82E+02	(	5.40E+01	)}	&	-6.33E+01	(	7.12E+01	)	&	F39	&	\bf{1.41E+06	(	8.00E+05	)}	&	3.77E+06	(	2.16E+06	)\\\hline
F12	&	\bf{-7.64E+01	(	4.85E+01	)}	&	4.40E+01	(	9.33E+01	)	&	F40	&	\bf{5.04E+04	(	2.64E+04	)}	&	7.04E+04	(	4.01E+04	)\\\hline
F13	&	\bf{1.30E+02	(	6.54E+01	)}	&	2.04E+02	(	8.10E+01	)	&	F41	&	\bf{7.08E+03	(	5.23E+03	)}	&	1.69E+04	(	1.37E+04	)\\\hline
F14	&	\bf{3.68E+03	(	4.56E+02	)}	&	4.27E+03	(	5.94E+02	)	&	F42	&	\bf{3.01E+04	(	2.25E+04	)}	&	3.59E+04	(	2.55E+04	)\\\hline
F15	&	\bf{3.88E+03	(	5.74E+02	)}	&	4.78E+03	(	7.29E+02	)	&	F43	&	3.01E+03	(	2.25E+02	)	&	\bf{3.00E+03	(	2.90E+02	)}\\\hline
F16	&	\bf{2.00E+02	(	1.13E-01	)}	&	2.01E+02	(	5.29E-01	)	&	F44	&	2.40E+03	(	2.44E+02	)	&	\bf{2.22E+03	(	1.91E+02	)}\\\hline
F17	&	\bf{5.28E+02	(	5.13E+01	)}	&	6.43E+02	(	1.03E+02	)	&	F45	&	\bf{1.23E+05	(	1.21E+05	)}	&	2.09E+05	(	2.49E+05	)\\\hline
F18	&	\bf{5.98E+02	(	2.85E+01	)}	&	7.64E+02	(	7.58E+01	)	&	F46	&	\bf{1.25E+05	(	6.31E+04	)}	&	2.66E+05	(	1.56E+05	)\\\hline
F19	&	\bf{5.04E+02	(	7.60E-01	)}	&	5.27E+02	(	1.13E+01	)	&	F47	&	\bf{2.67E+03	(	2.17E+02	)}	&	2.70E+03	(	2.23E+02	)\\\hline
F20	&	\bf{6.14E+02	(	2.94E-01	)}	&	6.15E+02	(	5.43E-01	)	&	F48	&	2.49E+03	(	3.14E+01	)	&	\bf{2.45E+03	(	2.66E+01	)}\\\hline
F21	&	1.02E+03	(	8.42E+01	)	&	1.02E+03	(	7.70E+01	)	&	F49	&	5.79E+03	(	2.04E+03	)	&	\bf{5.60E+03	(	2.60E+03	)}\\\hline
F22	&	\bf{5.36E+03	(	4.70E+02	)}	&	6.21E+03	(	7.47E+02	)	&	F50	&	3.26E+03	(	1.24E+02	)	&	\bf{2.87E+03	(	5.16E+01	)}\\\hline
F23	&	\bf{6.00E+03	(	7.84E+02	)}	&	6.39E+03	(	8.77E+02	)	&	F51	&	3.49E+03	(	9.56E+01	)	&	\bf{3.01E+03	(	4.49E+01	)}\\\hline
F24	&	1.31E+03	(	2.60E+01	)	&	\bf{1.30E+03	(	1.30E+01	)}	&	F52	&	\bf{2.89E+03	(	1.25E+01	)}	&	2.93E+03	(	2.13E+01	)\\\hline
F25	&	\bf{1.41E+03	(	1.03E+01	)}	&	1.42E+03	(	8.86E+00	)	&	F53	&	7.84E+03	(	1.80E+03	)	&	\bf{6.46E+03	(	6.20E+02	)}\\\hline
F26	&	\bf{1.46E+03	(	7.82E+01	)}	&	1.48E+03	(	8.69E+01	)	&	F54	&	3.85E+03	(	2.17E+02	)	&	\bf{3.34E+03	(	5.66E+01	)}\\\hline
F27	&	\bf{2.42E+03	(	1.07E+02	)}	&	2.58E+03	(	8.02E+01	)	&	F55	&	\bf{3.18E+03	(	3.60E+01	)}	&	3.23E+03	(	2.33E+01	)\\\hline
F28	&	\bf{2.03E+03	(	8.02E+02	)}	&	3.63E+03	(	1.45E+03	)	&	F56	&	\bf{4.40E+03	(	3.39E+02	)}	&	4.62E+03	(	3.21E+02	)\\\hline
	\multicolumn{2}{c}{  }  		         &                &             		F57	&	\bf{5.16E+05	(	2.90E+05	)}	&	1.54E+06	(	7.63E+05	)\\\cline{4-6}
\end{tabular}
\label{ruledata}
\end{table*}

\begin{table}[!htbp]
\centering
\caption{Results obtained by the Wilcoxon signed-rank test for ASBSO vs. BSO with $1/5$ Rule.}
\begin{tabular}{
c|c|c|c|c|c}\hline
 vs. & $R^{+}$ & $R^{-}$ & $p$-value &$\alpha$=0.05&$\alpha$=0.01 \\ \hline
BSO with $1/5$ Rule& 1282.0 & 371.0 &  2.22E-4 &YES&YES \\ \hline
\end{tabular}
\label{rule}
\end{table}

\begin{table*}[!htbp]
\centering
\caption{Experimental results  of using ASBSO and SFMS on CEC'13 and CEC'17 benchmark functions (F1-F57).}
\begin{tabular}{c|c|c|c|c|c}\hline
     \multicolumn{6}{c}{$D$=30}                                                                           \\\hline
  &ASBSO                                           & SFMS                               &   & ASBSO                       &SFMS \\\hline
  & Mean (Std Dev)                        & Mean (Std Dev)                   &     & Mean (Std Dev)             & Mean (Std Dev)  \\\hline
F1	&	-1.40E+03	(	1.98E-13	)	&	-1.40E+03	(	4.72E-13	)	&	F29	&	\bf{2.21E+03	(	2.00E+03	)}	&	3.13E+03	(	2.72E+03	)\\\hline
F2	&	\bf{1.54E+06	(	4.26E+05	)}	&	1.89E+06	(	4.58E+05	)	&	F30	&	\bf{3.95E+02	(	1.10E+02	)}	&	4.03E+02	(	1.18E+02	)\\\hline
F3	&	\bf{8.47E+07	(	8.64E+07	)}	&	1.16E+08	(	1.19E+08	)	&	F31	&	\bf{4.72E+02	(	2.92E+01	)}	&	4.98E+02	(	2.58E+01	)\\\hline
F4	&	\bf{5.08E+03	(	2.25E+03	)}	&	1.64E+04	(	4.84E+03	)	&	F32	&	\bf{6.86E+02	(	3.45E+01	)}	&	6.94E+02	(	3.34E+01	)\\\hline
F5	&	-1.00E+03	(	2.98E-03	)	&	-1.00E+03	(	2.41E-03	)	&	F33	&	\bf{6.51E+02	(	7.77E+00	)}	&	6.54E+02	(	7.62E+00	)\\\hline
F6	&	\bf{-8.64E+02	(	2.46E+01	)}	&	-8.61E+02	(	2.83E+01	)	&	F34	&	\bf{1.16E+03	(	9.94E+01	)}	&	1.17E+03	(	9.13E+01	)\\\hline
F7	&	\bf{-7.08E+02	(	3.90E+01	)}	&	-7.07E+02	(	3.46E+01	)	&	F35	&	\bf{9.41E+02	(	3.19E+01	)}	&	9.43E+02	(	2.14E+01	)\\\hline
F8	&	-6.79E+02	(	6.70E-02	)	&	-6.79E+02	(	9.63E-02	)	&	F36	&	\bf{3.93E+03	(	6.39E+02	)}	&	4.05E+03	(	7.25E+02	)\\\hline
F9	&	-5.71E+02	(	3.23E+00	)	&	-5.71E+02	(	3.42E+00	)	&	F37	&	\bf{5.20E+03	(	5.67E+02	)}	&	5.28E+03	(	6.97E+02	)\\\hline
F10	&	-5.00E+02	(	5.35E-02	)	&	-5.00E+02	(	8.43E-02	)	&	F38	&	1.23E+03	(	4.75E+01	)	&	1.23E+03	(	5.01E+01	)\\\hline
F11	&	\bf{-1.82E+02	(	5.40E+01	)}	&	2.34E+01	(	7.94E+01	)	&	F39	&	1.41E+06	(	8.00E+05	)	&	\bf{1.35E+06	(	7.99E+05	)}\\\hline
F12	&	\bf{-7.64E+01	(	4.85E+01	)}	&	1.73E+02	(	8.46E+01	)	&	F40	&	5.04E+04	(	2.64E+04	)	&	\bf{4.91E+04	(	2.42E+04	)}\\\hline
F13	&	\bf{1.30E+02	(	6.54E+01	)}	&	3.44E+02	(	7.70E+01	)	&	F41	&	\bf{7.08E+03	(	5.23E+03	)}	&	8.35E+03	(	7.67E+03	)\\\hline
F14	&	\bf{3.68E+03	(	4.56E+02	)}	&	3.88E+03	(	5.09E+02	)	&	F42	&	3.01E+04	(	2.25E+04	)	&	\bf{2.64E+04	(	1.34E+04	)}\\\hline
F15	&	\bf{3.88E+03	(	5.74E+02	)}	&	4.26E+03	(	5.32E+02	)	&	F43	&	\bf{3.01E+03	(	2.25E+02	)}	&	3.10E+03	(	3.99E+02	)\\\hline
F16	&	2.00E+02	(	1.13E-01	)	&	2.00E+02	(	1.99E-01	)	&	F44	&	2.40E+03	(	2.44E+02	)	&	\bf{2.39E+03	(	2.77E+02	)}\\\hline
F17	&	\bf{5.28E+02	(	5.13E+01	)}	&	5.54E+02	(	3.81E+01	)	&	F45	&	\bf{1.23E+05	(	1.21E+05	)}	&	1.35E+05	(	8.48E+04	)\\\hline
F18	&	\bf{5.98E+02	(	2.85E+01	)}	&	6.02E+02	(	3.11E+01	)	&	F46	&	\bf{1.25E+05	(	6.31E+04	)}	&	1.32E+05	(	5.56E+04	)\\\hline
F19	&	\bf{5.04E+02	(	7.60E-01	)}	&	5.06E+02	(	1.14E+00	)	&	F47	&	\bf{2.67E+03	(	2.17E+02	)}	&	2.73E+03	(	2.13E+02	)\\\hline
F20	&	6.14E+02	(	2.94E-01	)	&	6.14E+02	(	1.23E-01	)	&	F48	&	\bf{2.49E+03	(	3.14E+01	)}	&	2.50E+03	(	4.34E+01	)\\\hline
F21	&	1.02E+03	(	8.42E+01	)	&	1.02E+03	(	7.70E+01	)	&	F49	&	\bf{5.79E+03	(	2.04E+03	)}	&	6.11E+03	(	1.70E+03	)\\\hline
F22	&	\bf{5.36E+03	(	4.70E+02	)}	&	5.56E+03	(	7.10E+02	)	&	F50	&	\bf{3.26E+03	(	1.24E+02	)}	&	3.28E+03	(	1.01E+02	)\\\hline
F23	&	6.00E+03	(	7.84E+02	)	&	6.00E+03	(	6.64E+02	)	&	F51	&	\bf{3.49E+03	(	9.56E+01	)}	&	3.51E+03	(	1.25E+02	)\\\hline
F24	&	1.31E+03	(	2.60E+01	)	&	1.31E+03	(	2.27E+01	)	&	F52	&	2.89E+03	(	1.25E+01	)	&	2.89E+03	(	7.17E+00	)\\\hline
F25	&	\bf{1.41E+03	(	1.03E+01	)}	&	1.45E+03	(	1.68E+01	)	&	F53	&	7.84E+03	(	1.80E+03	)	&	\bf{7.48E+03	(	2.17E+03	)}\\\hline
F26	&	1.46E+03	(	7.82E+01	)	&	\bf{1.44E+03	(	7.15E+01	)}	&	F54	&	3.85E+03	(	2.17E+02	)	&	\bf{3.84E+03	(	1.96E+02	)}\\\hline
F27	&	\bf{2.42E+03	(	1.07E+02	)}	&	2.44E+03	(	1.30E+02	)	&	F55	&	\bf{3.18E+03	(	3.60E+01	)}	&	3.19E+03	(	4.05E+01	)\\\hline
F28	&	\bf{2.03E+03	(	8.02E+02	)}	&	5.66E+03	(	5.57E+02	)	&	F56	&	\bf{4.40E+03	(	3.39E+02	)}	&	4.47E+03	(	3.37E+02	)\\\hline
	\multicolumn{2}		{c}	{}		                                                  &                      &	F57	&	\bf{5.16E+05	(	2.90E+05	)}	&	5.31E+05	(	2.89E+05	)\\\cline{4-6}
\end{tabular}
\label{TMMdata}
\end{table*}

\begin{table}[!htb]
\centering
\caption{Results obtained by the Wilcoxon signed-rank test for IMS vs. SFMS.}
\begin{tabular}{
c|c|c|c|c|c}\hline
 vs. & $R^{+}$ & $R^{-}$ & $p$-value &$\alpha$=0.05&$\alpha$=0.01 \\ \hline
SFMS & 1343.5 & 309.5 &  2.0E-5&YES&YES \\ \hline
\end{tabular}
\label{tmm}
\end{table}

\subsection{ASBSO vs. previous BSO variants}
To further discuss the competitiveness of ASBSO, more comparisons between it and previous BSO variants should be executed. In this part, two BSO variants: BSO in objective space (BSOOS) \cite{shi2015brain} and global-best BSO (GBSO) \cite{el2017global} are tested on CEC'13 and 17 benchmark functions. The results are listed in Tables \ref{bsovar1} and \ref{bsovar2}.
\begin{table*}[!htbp]
\centering
\caption{Experimental results  of using ASBSO, BSOOS and GBSO on CEC'13 benchmark functions (F1-F28).}
\begin{tabular}{c|c|c|c}\hline
  &ASBSO                                           & BSOOS                                  & GBSO                    \\\hline
  & Mean (Std Dev)                        & Mean (Std Dev)                        & Mean (Std Dev)               \\\hline
F1	&	-1.40E+03	(	1.98E-13	)	&	\bf{-1.40E+03	(	1.64E-13	)}	&	-1.40E+03	(	1.06E-10	)\\\hline
F2	&	\bf{1.54E+06	(	4.26E+05	)}	&	1.67E+06	(	6.03E+05	)	&	2.08E+06	(	4.53E+05	)\\\hline
F3	&	\bf{8.47E+07	(	8.64E+07	)}	&	1.77E+08	(	2.89E+08	)	&	5.76E+08	(	7.05E+08	)\\\hline
F4	&	5.08E+03	(	2.25E+03	)	&	3.31E+04	(	9.12E+03	)	&	\bf{-1.04E+03	(	3.33E+01	)}\\\hline
F5	&	-1.00E+03	(	2.98E-03	)	&	-1.00E+03	(	1.72E-03	)	&	\bf{-1.00E+03	(	8.05E-04	)}\\\hline
F6	&	\bf{-8.64E+02	(	2.46E+01	)}	&	-8.62E+02	(	2.76E+01	)	&	-8.50E+02	(	3.16E+01	)\\\hline
F7	&	\bf{-7.08E+02	(	3.90E+01	)}	&	-6.82E+02	(	6.08E+01	)	&	-6.99E+02	(	2.76E+01	)\\\hline
F8	&	\bf{-6.79E+02	(	6.70E-02	)}	&	-6.79E+02	(	7.04E-02	)	&	-6.79E+02	(	7.42E-02	)\\\hline
F9	&	\bf{-5.71E+02	(	3.23E+00	)}	&	-5.69E+02	(	3.36E+00	)	&	-5.71E+02	(	3.40E+00	)\\\hline
F10	&	\bf{-5.00E+02	(	5.35E-02	)}	&	-5.00E+02	(	1.39E-01	)	&	-5.00E+02	(	9.27E-02	)\\\hline
F11	&	\bf{-1.82E+02	(	5.40E+01	)}	&	3.86E+01	(	7.84E+01	)	&	-1.48E+02	(	5.90E+01	)\\\hline
F12	&	\bf{-7.64E+01	(	4.85E+01	)}	&	1.42E+02	(	7.87E+01	)	&	-6.60E+01	(	6.99E+01	)\\\hline
F13	&	1.30E+02	(	6.54E+01	)	&	3.74E+02	(	1.10E+02	)	&	\bf{7.53E+01	(	5.32E+01	)}\\\hline
F14	&	\bf{3.68E+03	(	4.56E+02	)}	&	4.19E+03	(	5.57E+02	)	&	3.98E+03	(	5.46E+02	)\\\hline
F15	&	\bf{3.88E+03	(	5.74E+02	)}	&	4.23E+03	(	5.11E+02	)	&	4.22E+03	(	6.66E+02	)\\\hline
F16	&	2.00E+02	(	1.13E-01	)	&	\bf{2.00E+02	(	3.26E-02	)}	&	2.01E+02	(	2.31E-01	)\\\hline
F17	&	5.28E+02	(	5.13E+01	)	&	5.78E+02	(	4.43E+01	)	&	\bf{4.12E+02	(	2.11E+01	)}\\\hline
F18	&	5.98E+02	(	2.85E+01	)	&	5.99E+02	(	3.31E+01	)	&	\bf{5.09E+02	(	1.96E+01	)}\\\hline
F19	&	\bf{5.04E+02	(	7.60E-01	)}	&	5.05E+02	(	7.76E-01	)	&	5.07E+02	(	1.72E+00	)\\\hline
F20	&	\bf{6.14E+02	(	2.94E-01	)}	&	6.15E+02	(	3.16E-01	)	&	6.14E+02	(	6.14E+02	)\\\hline
F21	&	1.02E+03	(	8.42E+01	)	&	1.05E+03	(	8.37E+01	)	&	\bf{1.02E+03	(	7.70E+01	)}\\\hline
F22	&	\bf{5.36E+03	(	4.70E+02	)}	&	5.84E+03	(	9.16E+02	)	&	5.89E+03	(	8.82E+02	)\\\hline
F23	&	\bf{6.00E+03	(	7.84E+02	)}	&	6.30E+03	(	6.82E+02	)	&	6.21E+03	(	9.41E+02	)\\\hline
F24	&	1.31E+03	(	2.60E+01	)	&	1.35E+03	(	3.49E+01	)	&	\bf{1.29E+03	(	8.55E+00	)}\\\hline
F25	&	1.41E+03	(	1.03E+01	)	&	1.45E+03	(	2.30E+01	)	&	\bf{1.40E+03	(	1.40E+03	)}\\\hline
F26	&	1.46E+03	(	7.82E+01	)	&	1.54E+03	(	7.40E+01	)	&	\bf{1.45E+03	(	8.19E+01	)}\\\hline
F27	&	2.42E+03	(	1.07E+02	)	&	2.53E+03	(	1.08E+02	)	&	\bf{2.39E+03	(	1.07E+02	)}\\\hline
F28	&	\bf{2.03E+03	(	8.02E+02	)}	&	5.83E+03	(	5.81E+02	)	&	2.11E+03	(	1.00E+03	)\\\hline
\end{tabular}
\label{bsovar1}
\end{table*}

\begin{table*}[!htbp]
\centering
\caption{Experimental results  of using ASBSO, BSOOS and GBSO on CEC'17 benchmark functions (F29-F57).}
\begin{tabular}{c|c|c|c}\hline
  &ASBSO                                           & BSOOS                                  & GBSO                    \\\hline
  & Mean (Std Dev)                        & Mean (Std Dev)                        & Mean (Std Dev)               \\\hline
F29	&	2.21E+03	(	2.00E+03	)	&	\bf{1.96E+03	(	1.57E+03	)}	&	3.31E+03	(	4.07E+03	)\\\hline
F30	&	3.95E+02	(	1.10E+02	)	&	8.78E+03	(	3.04E+03	)	&	\bf{3.69E+02	(	5.23E+02	)}\\\hline
F31	&	4.72E+02	(	2.92E+01	)	&	\bf{4.66E+02	(	2.29E+01	)}	&	4.76E+02	(	1.19E+01	)\\\hline
F32	&	6.86E+02	(	3.45E+01	)	&	\bf{6.82E+02	(	2.88E+01	)}	&	6.92E+02	(	2.95E+01	)\\\hline
F33	&	6.51E+02	(	7.77E+00	)	&	6.50E+02	(	5.86E+00	)	&	\bf{6.46E+02	(	7.49E+00	)}\\\hline
F34	&	1.16E+03	(	9.94E+01	)	&	1.12E+03	(	6.90E+01	)	&	\bf{8.54E+02	(	4.02E+01	)}\\\hline
F35	&	9.41E+02	(	3.19E+01	)	&	\bf{9.37E+02	(	2.95E+01	)}	&	9.47E+02	(	3.06E+01	)\\\hline
F36	&	3.93E+03	(	6.39E+02	)	&	3.74E+03	(	4.78E+02	)	&	\bf{2.63E+03	(	1.02E+03	)}\\\hline
F37	&	\bf{5.20E+03	(	5.67E+02	)}	&	5.20E+03	(	8.49E+02	)	&	5.26E+03	(	5.85E+02	)\\\hline
F38	&	1.23E+03	(	4.75E+01	)	&	\bf{1.23E+03	(	4.18E+01	)}	&	1.26E+03	(	6.49E+01	)\\\hline
F39	&	\bf{1.41E+06	(	8.00E+05	)}	&	1.88E+06	(	1.30E+06	)	&	3.56E+06	(	2.70E+06	)\\\hline
F40	&	\bf{5.04E+04	(	2.64E+04	)}	&	5.84E+04	(	3.64E+04	)	&	8.29E+04	(	6.33E+04	)\\\hline
F41	&	7.08E+03	(	5.23E+03	)	&	9.92E+03	(	8.88E+03	)	&	\bf{6.09E+03	(	4.28E+03	)}\\\hline
F42	&	\bf{3.01E+04	(	2.25E+04	)}	&	3.23E+04	(	1.81E+04	)	&	4.94E+04	(	3.19E+04	)\\\hline
F43	&	3.01E+03	(	2.25E+02	)	&	3.10E+03	(	2.70E+02	)	&	\bf{2.95E+03	(	2.73E+02	)}\\\hline
F44	&	\bf{2.40E+03	(	2.44E+02	)}	&	2.42E+03	(	2.96E+02	)	&	2.41E+03	(	2.18E+02	)\\\hline
F45	&	\bf{1.23E+05	(	1.21E+05	)}	&	1.51E+05	(	1.21E+05	)	&	1.61E+05	(	1.20E+05	)\\\hline
F46	&	1.25E+05	(	6.31E+04	)	&	\bf{1.15E+05	(	4.55E+04	)}	&	4.66E+05	(	1.66E+05	)\\\hline
F47	&	\bf{2.67E+03	(	2.17E+02	)}	&	2.72E+03	(	2.10E+02	)	&	2.70E+03	(	1.20E+02	)\\\hline
F48	&	\bf{2.49E+03	(	3.14E+01	)}	&	2.51E+03	(	3.67E+01	)	&	2.50E+03	(	2.14E+01	)\\\hline
F49	&	5.79E+03	(	2.04E+03	)	&	6.42E+03	(	1.54E+03	)	&	\bf{4.17E+03	(	2.19E+03	)}\\\hline
F50	&	3.26E+03	(	1.24E+02	)	&	3.31E+03	(	9.91E+01	)	&	\bf{3.03E+03	(	9.78E+01	)}\\\hline
F51	&	3.49E+03	(	9.56E+01	)	&	3.47E+03	(	2.09E+02	)	&	\bf{3.14E+03	(	1.04E+02	)}\\\hline
F52	&	2.89E+03	(	1.25E+01	)	&	\bf{2.88E+03	(	8.40E+00	)}	&	2.90E+03	(	2.81E+01	)\\\hline
F53	&	7.84E+03	(	1.80E+03	)	&	7.65E+03	(	1.89E+03	)	&	\bf{5.99E+03	(	1.60E+03	)}\\\hline
F54	&	3.85E+03	(	2.17E+02	)	&	3.86E+03	(	2.64E+02	)	&	\bf{3.25E+03	(	8.17E+01	)}\\\hline
F55	&	\bf{3.18E+03	(	3.60E+01	)}	&	3.21E+03	(	1.35E+01	)	&	3.22E+03	(	2.58E+01	)\\\hline
F56	&	4.40E+03	(	3.39E+02	)	&	\bf{4.37E+03	(	2.71E+02	)}	&	4.41E+03	(	3.32E+02	)\\\hline
F57	&	\bf{5.16E+05	(	2.90E+05	)}	&	7.73E+05	(	4.79E+05	)	&	1.36E+06	(	7.50E+05	)\\\hline
\end{tabular}
\label{bsovar2}
\end{table*}

From the results, ASBSO shows a great advantage comparing with BSOOS, and can be competitive with GBSO. Although the $p$-value for ASBSO vs. GBSO is not less than 0.05, ASBSO still obtains a greater $R^{+}$ value, which indicates that it has a better overall performance than GBSO on total 57 test functions. Moreover, GBSO adopts multiple modifications, i.e., fitness-based grouping, per-variable updates, the global-best update and the re-initialization step, but ASBSO using fewer modifications obtains competitive results, which could be regarded as a successful variant of BSO.

\begin{table}[!htp]
\centering
\caption{Results obtained by the Wilcoxon test for algorithm ASBSO vs. BSOOS and GBSO.}
\begin{tabular}{cccccc}\hline
 Algorithms & $R^{+}$ & $R^{-}$ & $p$-value  &$\alpha$=0.05&$\alpha$=0.01 \\ \hline
 ASBSO vs.                                                             \\ \hline
BSOOS & 1292.5 & 303.5 &  0.000031    & YES & YES\\ \hline
GBSO & 961.0 & 635.0 &  0.163962   & NO & NO\\ \hline
\end{tabular}
\label{bsovar3}
\end{table}

\section{Discussion}
\label{section5}
As shown fully detailed in Section \ref{section4}, our proposed ASBSO outperforms traditional BSO and other meta-heuristic optimization algorithms. Especially in comparison with MABC and CGSA-M which also implement memory-based selection mechanism, ASBSO obtains much better results in solution accuracy. It is interpreted in Section \ref{section3} that ASBSO has two main novelties: first, it adapts several step length update methods to deal with different situations; second, these methods are adaptively selected via a new memory storing mechanism. In this section, we will further discuss the effectiveness of these two modifications by comparing them with the classical $1/5$ success rule used in evolutionary strategy (ES) \cite{rechenberg1978evolutionsstrategien} and SFMS used in \cite{song2017multiple, qin2009differential}, respectively. These tests are executed at $D=30$ with maximum number of function evaluation equals 10000$D$ for 30 runs.

\subsection{Comparison with $1/5$ Success Rule}
$1/5$ success rule is a parameter adaptive strategy proposed by Rechenberg \cite{rechenberg1978evolutionsstrategien} which is used to adjust deviation $\delta$ in order to make mutational step size be dynamically adapted according to the search performance.

The offspring generation equation can be exhibited as follow:

\begin{equation}
X_{offspring}=X+N(0, \delta(t))
\label{es1}
\end{equation}
where $X$ is the parent and $X_{offspring}$ is the offspring. It is generated by adding a Gaussian noise $N(0, \delta(t))$ of which  mean value equals 0 and deviation $\delta(t)$ changes according to iteration $t$.

Its variation equation can be shown as:
\begin{equation}
\delta(t+1)=\left\{ \begin{array}{lll}\frac{\delta(t)}{r} &  if & s_r >0.2 \\
                                                                  \delta(t)*r  & if & s_r <0.2 \\
                                                                  \delta(t)    &  if  &s_r=0.2 \end{array}\right.
\label{es2}
\end{equation}
where $r$ is a scale factor that is usually set in interval $[0.85,0.99]$, and $s_r$ is a success rate to represent the rate that mutation procedure successfully generates a better offspring in a certain period. If the success rate $s_r$ is larger than 0.2, deviation $\delta$ will increase; in the opposite, if $s_r$ is smaller than 0.2, $\delta$ will decrease. As an adaptive mechanism, it makes algorithm can adjust its search radius to be suitable for specific problems and different search periods. Not only in ES, but also in some other newly proposed algorithms, such as negatively correlated search proposed by Tang et al. \cite{tang2016negatively}, $1/5$ success rule has exhibited a great performance in search ability. Thus, we combine BSO with $1/5$ success rule to conduct a contrast experiment to assess the effectiveness of ASBSO.

Table \ref{ruledata} lists the experimental results between ASBSO and BSO with $1/5$ success rule on 57 test functions. It is obvious that although $1/5$ success rule can obtain better solutions on a few problems, ASBSO still dominates most number of the problems. Table \ref{rule} shows the Wilcoxon statistical analysis result between ASBSO and BSO with $1/5$ success rule, where ASBSO is the control algorithm. $p$-value that is smaller than significant level $\alpha=0.01$ demonstrates that the multiple step length update method proposed in ASBSO can provide more adaptive and suitable search mechanisms than the $1/5$ success rule to be applied to various problems.

\subsection{IMS vs. SFMS}
The second modification of the proposed method is that a new memory storing mechanism IMS replaces the traditional memory mechanism (SFMS). Both mechanisms are introduced in Section \ref{section3} and it is necessary to discuss whether the former can provide a better search efficiency than the latter. Hence, a comparison between ASBSO and the BSO with adaptive step length based on SFMS is conducted and the results are listed in Table \ref{TMMdata}. Visually, ASBSO maintains most better results especially on CEC'13. Table \ref{tmm} also can prove that IMS is significantly better than SFMS.

\subsection{Computational Complexity}
ASBSO has shown a superior ability for a majority of benchmark functions. In this subsection, we calculate its computational time complexity together with BSO's.

The time complexity in each procedure of BSO is described as follows:

(1) In BSO, the time complexity for initializing is $O(N)$ where $N$ is the population size.

(2) Evaluating the fitness of population is $O(N)$.

(3) Using K-means to divide the population into $c$ clusters needs $O(cN^2)$.

(4) The process of individual selection and step length generation both cost $O(N^2)$.

(5) The generation of new individuals and the fitness calculation need $O(N^2)$, respectively.

Thus, the overall time complexity of BSO is

\begin{multline}
O(N)+O(N)+O(cN^2)+O(N^2)+O(N^2)\\
=2O(N^2)+O(cN^2)+2O(N)
\end{multline}

To be simplified, its overall time complexity is $O(N^2)$.

ASBSO is modified based on BSO. Its procedure is shown as:

(1) The initialization needs $O(N)$.

(2) Evaluating the fitness of population is $O(N)$.

(3) Using K-means to divide the population into $c$ clusters needs $O(cN^2)$.

(4) Generate multiple step lengths needs $O(4N^2)$.

(5) The memory selection costs $O(N)$.

(6) The generation of new individuals and the fitness calculation need $O(N^2)$, respectively.

Thus, the overall time complexity of ASBSO is

\begin{multline}
O(N)+O(N)+O(cN^2)+O(4N^2)+O(N)+O(N^2)\\
=O(cN^2)+O(4N^2)+O(N^2)+3O(N)
\end{multline}

The overall time complexity of ASBSO can be seen as $O(N^2)$. The main differences between ASBSO and BSO are in Steps (4) and (5). As ASBSO applies multiple step length strategies, it costs $O(4N^2)$ which is greater than $O(N^2)$ of BSO, and the memory selection needs $O(N)$. Thus, ASBSO and BSO have the same time complexity, which indicates that both are competitive in computational efficiency.

\section{Conclusion}
\label{section6}
In this paper, an adaptive step length mechanism based on memory is proposed for BSO, namely ASBSO. It applies multiple step length generation strategies and a new memory mechanism in aim to generate better individuals for different search periods and problems. The strategies with different step lengths are produced by using four different scale parameters and they are selected based on a memory structure in each iteration. Different from the conventional memory mechanism, the proposed memory structure method is created to record the improvement value in fitness obtained by each strategy. By implementing this, the strategy which can increase solution quality substantially has a higher possibility to be selected compared with the original one which can similarly success while obtains only a little improvement. The performance of ASBSO has been tested by using CEC'13 and CEC'17 benchmark function suits (57 functions in total) which include different characteristics. Some well-known optimization algorithms also have been added into comparison. Experimental and statistical results show that the proposed ASBSO can succeed in improving the performance of BSO in terms of global search ability, convergence speed, robustness and solution quality. Moreover, some real-world problems in CEC'11 are introduced to present the application value of ASBSO. These results can encourage our future research into self-adaptive search mechanism. Furthermore, this will broaden our perspective of BSO for dynamic and multiobjective optimization.

\bibliographystyle{IEEEtran}
\bibliography{mybibfile}

\end{document}